%% file: main.tex
\documentclass[sigconf]{acmart}

\def\orcidID#1{\unskip$^{[#1]}$}

\setcopyright{none}

\usepackage{bbm}

\usepackage{graphicx}
\graphicspath{{figs/}}

\usepackage{subcaption}
\usepackage{listings}
\usepackage{todonotes}

\usepackage{enumitem}

\usepackage{fancyvrb}
\usepackage{fvextra}

\usepackage{booktabs}
\usepackage{multirow}

\usepackage{cleveref}

\crefname{todo}{todo}{todos}
\Crefname{todo}{Todo}{Todos}

\usepackage{siunitx} %

\usepackage{amsmath}

\usepackage{xcolor}
\usepackage{soul}

\usepackage{titlesec}

\titleclass{\subsubsubsection}{straight}[\subsection]

\newcounter{subsubsubsection}[subsubsection]
\renewcommand\thesubsubsubsection{\thesubsubsection.\arabic{subsubsubsection}}

\titleformat{\subsubsubsection}
  {\normalfont\normalsize\bfseries}{\thesubsubsubsection}{1em}{}
\titlespacing*{\subsubsubsection}
{0pt}{3.25ex plus 1ex minus .2ex}{1.5ex plus .2ex}

\makeatletter
\renewcommand\paragraph{\@startsection{paragraph}{5}{\z@}%
  {3.25ex \@plus1ex \@minus.2ex}%
  {-1em}%
  {\normalfont\normalsize\bfseries}}
\renewcommand\subparagraph{\@startsection{subparagraph}{6}{\parindent}%
  {3.25ex \@plus1ex \@minus .2ex}%
  {-1em}%
  {\normalfont\normalsize\bfseries}}
\def\toclevel@subsubsubsection{4}
\def\toclevel@paragraph{5}
\def\toclevel@paragraph{6}
\def\l@subsubsubsection{\@dottedtocline{4}{7em}{4em}}
\def\l@paragraph{\@dottedtocline{5}{10em}{5em}}
\def\l@subparagraph{\@dottedtocline{6}{14em}{6em}}
\makeatother

\acmConference{\ }{\ }{\ }
\acmISBN{}
\acmDOI{}

\begin{document}

\begin{abstract}
\input{abstract}
\end{abstract}

\title{Testing the Depth of ChatGPT's Comprehension via Cross-Modal Tasks Based on ASCII-Art}
\subtitle{GPT3.5's Abilities in Regard to Recognizing and Generating ASCII-Art Are Not Totally Lacking}
\author{David Bayani\orcidID{0000-0001-5811-6792}}
\email{dcbayani@alumni.cmu.edu}
\email{david.bayani@inpleo.com}
\orcid{000-0001-5811-6792}

\maketitle

\section{Introduction}

\input{introduction}

\section{Related Work}

\input{related_work}

\section{ASCII-Art Recognition Experiments}

\input{recognitionExperiments}

\section{ASCII-Art Generation Experiments}

\input{generationExperiments}

\section{Conclusion}

\input{conclusion}

\section{Future Work}

\input{futureWork}

\clearpage

\bibliographystyle{ACM-Reference-Format}
\bibliography{Bibliography-File}

\appendix

\input{appendix}

\end{document}

%% file: abstract.tex
Over the eight months since its release, ChatGPT and its underlying model, GPT3.5, have
garnered massive attention, due to their potent mix of capability and accessibility. 
While a niche-industry of papers have emerged examining the scope of capabilities these 
models possess, the information fed to and extracted from these networks has been either
natural language text or stylized, code-like language.
Drawing inspiration from the prowess we expect a truly human-level intelligent agent to have 
across multiple signal modalities, 
in this work we examine GPT3.5's aptitude for visual tasks, where the inputs 
feature content provided as ASCII-art without overt distillation into a lingual summary. 
We conduct experiments analyzing the model's performance on image recognition tasks after 
various transforms typical 
in visual settings, trials investigating knowledge of image parts, and tasks covering image generation.

%% file: introduction.tex
Large language models (LLMs) have taken the world by storm over the last several
years, 
grabbing headlines in virtually every media outlet (examples spanning television, radio and newspapers include \cite{NBCNightlyNews_d22M12y2022,NPR_d10M1y2023,NewYorkTimes_d15M4y2022}).

As an  instance of particular note, ChatGPT has garnered mass attention and rapid adoption
since its public release in November 2022. ChatGPT builds off of version 3.5 of the Generative Pre-trained Transformer model family developed by OpenAI, a child whose lineage has
been marked by one massive step after another in regard to the size of LLM networks and their training data.
In light of its ground-breaking performance across an array of NLP tasks, 
ChatGPT/GPT3.5 
is already finding use in industry for assistance in travel-planning, language learning, internet search, and workplace instant-messaging \cite{forbes_d30M5y2023}. 
In addition to 
recognition in the commercial world, it has gained
the rare honor of becoming a machine-learning model 
that is a household-name --- an occurrence that 
appears to 
coincide with growing concerns over 
the ramifications of modern artificial intelligence on education and plagiarism identification \cite{CBSMinnesota_d19M1y2023}.
Fueled by the model's unprecedented popularity, accessibility, and power, 
a niche-industry of papers attempting to 
rigorously 
investigate the abilities of ChatGPT --- and the GPT3/GPT3.5 family underlying it more broadly --- have 
materialized in short order.

Drawing from the fact that
these models having been designed for primarily natural language
tasks, many efforts have focused on language-centric or language-exclusive tasks \cite{liu2023summary}.
As has now been illustrated, however, ChatGPT appears capable of outputting
reasonable code on request and generating other content belonging to certain 
synthetic, technically-oriented languages, such as those prevalent in 
early-stage engineering courses \cite{deshpande2023analyzing,sadik2023analysis}.

In this paper, we explore GPT3.5's abilities in an under-trodden direction: 
we explore how well can the model
"see" and "draw". That is, while the model may display spatial reasoning once
visual stimuli has been distilled into symbolic descriptions, 
it is not nearly as well established to what degree it can actually recognize and generate visual patterns, relations, and depictions
absent any substantial modification or additional pre-/post-processing by other foundation models that were explicitly crafted for multi-modal settings.
Our 
vehicle for delivering visual content during this analysis is ASCII-art \cite{britanicca_d19M6y2014}.

Given the at-times eyebrow-raising discourse surrounding these recent technological advances --- 
comparisons being made between the latest generation of chatbot and artificial general intelligence, questions over the
continued adequacy of the Turing test as an AI-benchmark, concerns over the displacement of skilled professionals in the workforce 
and so forth --- examining the extent of such system's
aptitude outside of the theater for which it was specifically optimized does not strike us a meritless line of inquiry. 
Indeed, 
we believe it a worthy venture to determine how well a model capable of holding its own on the BAR-exam against 
competent adults \cite{reutersGPTFamilyAndBarExam} could compete with a toddler when it comes to the most basic activity outside of prose.

%% file: related_work.tex
\subsection{\small VQA, Image Generation, Robotics, et al.: The Most Common Way to Let ChatGPT See is Giving It a Seeing-Eye Dog}

Most work relating to ChatGPT (or
the GPT-family of models  that predate the vision-equipped GPT4) have considered
problems within the canon of natural language and writing tasks \cite{zhang2023small,liu2023summary,zhong2023agieval}.
This characterization serves more to connote the typical deployment of the text-based medium 
used to exchange information than to substantially narrow the type of information
exchanged; tasks ranging from writing poetry \cite{gptAndPoetry_cushman}, to programming \cite{sadik2023analysis},
to room navigation problems \cite{joublin2023glimpse} have been conveyed to and addressed by ChatGPT via verbal description.
As pointed out in \cite{liu2023summary}, ChatGPT's diverse capabilities and easy accessibility have
helped fuel a deluge of papers exploring the systems potential and limitations. 
Within this space, most relevant to us are efforts that treat ChatGPT's spatial reasoning,
as well as the works exploring integration of ChatGPT in a multi-component process-pipelines geared toward 
text-based image recognition, manipulation, or
generation.
 
Examining ChatGPT's mastery of an introductory-level electrical engineering course,
\cite{deshpande2023analyzing} observe that while the model does quite well at language-only 
tasks, "predictably, as a text-only tool, it cannot handle questions with diagrams or figures, 
nor can it generate diagrams and figures." While conveying the take-away from their evidence
correctly for the most-part, the quoted summary is muddied by the author's earlier observation
that ChatGPT appears to, at times, attempt drawing ASCII-art circuits, albeit in not a totally legible fashion.  

Quoting the summary provide by \cite{zhang2023small} in regard to findings collected by \cite{borji2023categorical},
``ChatGPT lacks a `world model' to perform spatial, temporal, or physical inferences, or
to predict and explain human behaviors and psychological processes [...], and is also limited in mathematics and
arithmetic, unable to solve difficult mathematical problems or riddles, or even possibly get inaccurate results in some
simple computation tasks''. For the most part, the ``spatial''
tasks covered by \cite{borji2023categorical} where performed by describing verbally a
situation and asking the model to make an inference based on the facts conveyed; 
while a few examples from the piece include ASCII-art,\footnote{Figures 6 and 18 in that work, \cite{borji2023categorical}.}
both are figures output by the model and are of a generic enough nature to have been conceivably
(if not most probably) taken verbatim from training data.

 Complicating the picture, 
in an intriguing demonstration of what we will characterize as  ChatGPT's apparent ``algebra''-like spatial reasoning capabilities, \cite{ye2023improved} used the LLM to mediate between a human and robot collaborator.
In experiments, the language model received natural language commands from a human undertaking a construction
task and either requested further clarification prior to proceeding, or translated the command into 
simplified, procedural steps for execution by a robotic arm controller - for example, 
$$Grab [driller] Move [0.2, 0, 1] Drop [driller]$$
While the robotic arm's planner and motor controls were the components primarily responsible for
elaborating basic commands into environmental actions,
ChatGPT needed to model the 
situation sufficiently well to be able to raise questions when
appropriate.
While the authors demonstrate a working vision-compatible pipeline that contains the LLM as a component, 
even under the most
expansive interpretation of its role in \cite{ye2023improved}, ChatGPT deals with space as distilled into
verbal descriptions, and the actual degree to which it needs to 
model the 3D world beyond maintaining a vocabulary of location and object names is questionable.%

Investigations carried out by \cite{joublin2023glimpse} contained a substantial focus on
planning and logical reasoning in a spatial setting. Among their trials with ChatGPT were 
intuitive physics experiments (e.g., which way a shadow would be cast considering light 
conditions),  one dimensional object ordering tasks, two dimensional box placement 
questionnaires, simulated robotics exploration (navigating an apartment and
searching a room for a ball), and simulated robotics task completion (setting a table for a 
meal). Limiting the applicability of these efforts to our focuses, however, is the fact that
all "spatial" or "visual" interactions were conveyed in high-level terms. For example,
navigating the apartment consisted of the human telling ChatGPT which room it was in and which doors
it had available to choose from, then describing the new room and doors options after the AI
selected a choice. On the whole, the results by \cite{joublin2023glimpse} seem to demonstrate
that ChatGPT is at least able to track objects present in a context and, to some degree, 
reason about space "algebraically" if not not "geometrically".
Using a text-world game as crucible for experiments,
work by \cite{chen2023introspective} investigating the network's  contextual decision making capabilities
is similarly relevant to us, although also having similar deficits in regards to addressing questions central
to our focus.

There have been attempts to integrate recent GPT-family models into VQA.
For example, \cite{yang2022empirical} used image-to-text networks to produce 
descriptions of the target image, user query, and several other similar in-context question-and-answer
examples which are then fed into GPT-3 with the user's original input in order to generate
a response.
In \cite{bongini2022gpt3}, the authors rely on GPT-3 to provide descriptions of paintings given their names,
which is then
fed forward to a question-answering model to reply to queries.
Their approach in fact incorporates \textit{no} direct ingestion of the 
image, instead conducting all activity in text and relying on 
``the memorization capabilities of GPT-3, which at training time has observed millions of tokens regarding
domain-specific knowledge.''
In contrast, \cite{si2023combo} leveraged a variety of knowledge sources --- spanning from %
explicit (e.g., Wikipedia) to implicit (e.g., GPT-3) resources --- that digested information
in various modalities prior to forming a combined embedding which is fed into a final decoder area.
While the overall pipeline constructed by \cite{si2023combo}
is more sophisticated and
does attempt to handle visual information without it all being converted to text first, in respect to 
garnering useful information from GPT-3, the author's approach is not entirely unlike \cite{yang2022empirical},
presenting the LLM with verbal descriptors of the scene, the query, and in-context examples 
in its prompt; that is, while parts of their work treat images as first-class citizens to process, 
GPT-3 is presented a typical, natural-language-text-input-natural-language-text-output setting.
These works are not alone in their attempts to benefit from ChatGPT's obvious competences outside of 
visual modalities for tasks in that regime;
indeed, works such as \cite{chalvatzaki2023learning,tiong2023plugandplay,li2023lmeye,huang2023instruct2act,mu2023embodiedgpt,srivastava2023generating}
also explore
ingesting pictures for tasks that have GPT3.5-or-prior in the loop, 
\cite{yang2023mmreact,maddigan2023chat2vis,nanwani2023instancelevel,qin2023instructvid2vid}\footnote{Also in this list is the commercial-venture Botto (\url{https://docs.botto.com/details/bottos-art-engine}), a largely-automated ``artist'' that uses GPT3 to
    feed prompts to other text-to-image generative models.} explore including the component in pipelines that \textit{produce} images,
and
outings like \cite{shen2023hugginggpt,wu2023visual} consider settings where 
visual-content could play either input or output roles.\footnote{Given the language-base input and output encodings used, some works, such as the robotics-centered venture
in \cite{huang2023instruct2act,mu2023embodiedgpt}, could be debatably considered to output encoded "visual" artifacts in a similar fashion as the LLMs play in a number of the image-generation works cited.
In this case, we err towards where explicit image/video input/output is present to guide
the fuzzy grouping presented in this sentence.}
The diversity of implementation-specifics notwithstanding across this set, the takeaway is largely the same:
these works either (1) summarize context verbally or in a human-readable data structure or (2) modify the models in question
to explicitly include visual knowledge, often  
coupling this with additional training of parts that are woven intimately into the LLMs.

Another growing (sub-)sub-area of tangential pertinence to us is the use of LLMs, particularly of the GPT-family, 
for graph analysis, understanding, and visualization \cite{zhang2023graphtoolformer,wang2023language,guo2023gpt4graph,shi2023chatgraph,zhu2023llms,dibartolomeo2023ask};
conceivably, experimenters might attempt to use character-art or some similar method to directly feed illustrations of 
graph-structures into the network, depictions similar to those found in a computer science 101 class.
Our imagination
notwithstanding, we find that the gaps existent in the preceding paragraph's references apply to this collection as well, insofar as 
how the LLMs treat any "visual" content.

We would be remiss to fail noting that the latest iteration of the GPT model family, GPT-4 \cite{openai2023gpt4},
accepts visual input and is able to answer questions referencing them. In addition to simply being a different 
model than the one in our current locus of analysis, we do not scrutinize GPT-4's vision capabilities
since we consider analysis of an architecture
specifically built and trained to operate with that modality to be beside the point of our present effort.
On that topic, we comment further as to our rationale and this paper's inspiration in \cref{subsec:relatedWork:inspiration}.

\subsection{\small GPT's ``Algebraic'' Handling of Spatial and Visual Information in Most Prior Work}

Elaborating an earlier comment: if we imagine rotations of an equilateral triangle about its center, 
we can determine the results of rotating clockwise by $\pi$ radians followed by $-\pi$ radians either
by computing how the points of the triangle would be transformed in each step of the process ---
in the spirit of a "geometric" understanding, as we mean it --- or by reasoning about the rotations
from the "algebraic" perspective of group actions, where the composition of an action and its inverse
result in the identity. Observe that in this example, it \textit{might} be argued that the algebra truly captures
part of what it means to "understand" space; however, even if that is true, the depth of penetration is 
limited --- while I can tell you specific occurrences of algebraic facts by tracking points "geometrically",
I cannot tell you much about points if I treat space like a free group.\footnote{A free group is 
essentially a group with no assumptions on it other than what is required by the group axioms.
As such, we know $a a^{-1} = I$ but we don't know $a b a^{-1} b^{-1} = I$, even if in the latter case
it corresponds to moving $y$ units left, rotating $x$ degrees clockwise, moving $-y$ using right, 
and rotating $-x$ degrees counterclockwise in the two dimensional Euclidean group. One can consider
settings with additional properties not assumed by free groups - such as abelianness - but  
even in those setting, there is an intuitive sense where one must step down the abstraction level, provide
more detail, and do additional nitty-gritty work in order to judge what $a b$ means in respect to 
changing a collections of 2D points. }
As a theme throughout the whole of serious 
VQA and image generation works that attempt
to leverage ChatGPT in some way, 
the model's adeptness in comprehending visual content prior to it being abstracted into a verbal description is
largely untreated.
\footnote{We say "largely" since on rare occasion works have displayed rudimentary 
drawings that, under a liberal interpretation, might be considered in a grey-area. For example, in the case of the one dimensional
spatial ordering tasks displayed in \cite{joublin2023glimpse}, ChatGPT produced - without explicit prompting to do so - a rudimentary line drawing
as part of its answer, a fact which blurs the line ever so slightly.}

While it may be tempting to believe that effective manipulation or generation of graphics via specifying
program commands demonstrates competence in vision to a comparable degree as, say, analysis of a raw bitmap
image, it may be prudent to pause before one jumps to that conclusion --- if nothing else than in light of the popular 
ridicule of older generations of artificial intelligence work that made pervasive use of symbolic methods for vision tasks
(for instance, see \cite{DBLP:journals/ai/Brooks91,moravec1993universal,sutton2019bitter} for a
couple
critical takes\footnote{Related to these sources, one may want to become acquainted with the
so-called \href{https://en.wikipedia.org/wiki/Moravec\%27s_paradox}{\color{blue}"Moravec Paradox"},
hinted-at in the short foray cited and treated more thoroughly in the
author's earlier book, \cite{moravec1988mind}. A further digression about this general subject is in \cref{appendix:sec:symbolicVision}.}).

\subsection{\small Works Closest to Our Own}
\label{subsec:relatedWorks.closest}

Outside of one-off curiosities which resonate with aspects of this work's spirit and are found occasionally among informal, public communications,
the substantive works closest to ours
are \cite{Todd_2023}, \cite{wang2023bot}, and \cite{DBLP:journals/corr/abs-2306-07195}.

In exploring prior work, we uncovered that there is a certain degree of "folk-knowledge" about
ChatGPT's drawing abilities --- for instance \cite{internet_folk_wisdom_1,internet_folk_wisdom_2,internet_folk_wisdom_3,internet_folk_wisdom_4,internet_folk_wisdom_5,internet_folk_wisdom_6,internet_folk_wisdom_7,internet_folk_wisdom_8,internet_folk_wisdom_9,internet_folk_wisdom_10,internet_folk_wisdom_11,internet_folk_wisdom_12,internet_folk_wisdom_13,internet_folk_wisdom_14,internet_folk_wisdom_15,internet_folk_wisdom_16,internet_folk_wisdom_17,internet_folk_wisdom_18}. To the extent that we have seen, however, 
exchanges in this category that focus on ASCII-art directly --- as opposed to code to generate diagrams, etc. ---  were
mostly sporadic acts of passing curiosity, not systematic, extensive, or particularly deep explorations.
A theme that is present in these folk-studies is the appearance of ASCII-art of reasonable quality, but
occurring at inappropriate times in respect to the prompts; we suspect that these occurrences again reflect
some degree of memorization by the network, namely that, when prompted, it can return something scraped from
the web that is known to be ASCII-art, but in general fails to capture what the character-drawing is meant
to depict. There may be exceptions to this for particularly common depictions which have received explicit
human commentary --- smiley faces, small cats, various emoji-like strings, etc.
Part of our investigation is to determine how well-informed and flexible 
ChatGPT's visual and spatial faculties are, in particular getting a sense of how competently they extend beyond
shallow memorization and if the network can distinguish between ASCII-art illustrations in a non-trivial sense.

In \cite{Todd_2023}, the authors explore the use of GPT2, and to a limited extent GPT3 (not GPT3.5 or higher), to generate
small two-dimensional game levels, where the output of the model is a specification in ASCII-art.
Outside of the generation of ASCII-art with a related family of model as ours, similarities largely
end. Unlike our efforts to essentially judge knowledge already present in GPT3.5, \cite{Todd_2023} 
perform additional pre-training for their specific game-generation tasks, in addition to 
variously adopting 
specially curtailed tokenizers. In short, they take active steps to attempt reworking 
the basic model to perform well for their task - which is perfectly reasonable given their primary goal
of level generation. They do not evaluate the ability of their models to perform question answering on 
ASCII-art input, nor do they rigorously investigate the set of considerations that typically are associated
with computer vision, such as toleration to noise, translation invariance, or even recognition.

Under the impetus of differentiating ChatGPT-generated content from that of humans,
\cite{wang2023bot} curated categories of questions that emphasized the areas where LLMs' 
aptitude most differed from that of a human, either by under-performing or out-performing.
Among the eight tests considered, identification of ASCII-art provided by the interviewer
was one, with a patent gap between human and ChatGPT performance being borne out in the results --- 94\% and
8\% accuracy respectively on 50 unique drawings sampled from a catalog.
In addition to the limited show-verbatim-and-describe nature of these ASCII-art trials, 
worthy of note is the fact that the samples were gathered from a publicly available
website which was on the web for at least four years prior to 
ChatGPT's release, the \href{www.asciiart.eu}{\color{blue}ASCII Art Archive},\footnote{
A snapshot of content from March 2018 can be seen \href{https://web.archive.org/web/20180305160309/https://www.asciiart.eu/}{\color{blue}here}.
Older 
versions of the website at a compatible URL seem to be listed on the Internet Archive \href{https://web.archive.org/web/20091225193406/http://asciiart.eu:80/}{\color{blue}here}, however this looks to be a page describing the concept, not providing cataloging examples. } risking appearance in ChatGPT's training set;
moreover, the examples from the catalog used might have already been in general circulation 
before being collected together.
While 8\% accuracy is not astounding performance, it is not nothing, though questions remain
about how much of that is from memorization versus more 
general recognition.

Most similar in spirit and relation is the very recently released work
\cite{DBLP:journals/corr/abs-2306-07195}.
Like us, their concern is to examine existing capabilities of members of OpenAI's Generative Pre-Trained model 
family, GPT3.5 and GPT4 in their case, using a series of prompts without additional training or system modifications.  Figure 6, 14, and 18  in their work are latex output, which fall into the category of code and visualization-via-code output that we 
discussed earlier.
Figures 16 and 17, however, are concerned with leveraging rudimentary ASCII-art as another avenue to explore recursive output.
There is a grey line whether this is a "visual" task or essentially an algebraic computation.
The authors display slight caution when evaluating the LLM's performance in generating
certain character-level depictions --- weighing the possibility that
a subset of the ASCII-art patterns displayed where memorized from the training data, especially
those that were highly structured and resemble famous examples ---
 but   point-out that the more exotic productions among the figures 
 are subject
to this concern less.\footnote{Minor connections following from this are listed in \cref{appendix:subsec:twoSmallObsAndConnections}.} 

\subsection{\small The Inspiration for this Work}
\label{subsec:relatedWork:inspiration}

The conceptual seed for this endeavor came from a news-item circa June 2022, when an employee at Google
claimed that LaMDA, one of the company's large language models, was sentient, as judged from interactions he
had had with it \cite{washingtonPost_d11M6y2023}.
A natural question that comes to the mind of any enthusiast is what they, if given the chance, would 
query the model with. The notion of being the inquisitor in a Turing test, facing-down a machine parading 
as human, looms large in the background. %
This sprung to mind posing questions about ASCII-art since
a demonstration that LaMDA failed to have 
rudimentary abilities in modalities outside text would at least
temper the illusion that the computer had ascended to the full anthropomorphic heights.\footnote{This said, as reported in \cite{washingtonPost_d11M6y2023} and
legendarily the original ELIZA, those who are already convinced may find routes of logic around 
evidence that counters their belief. 
We make no comment here in regard to holding beliefs that run counter to orthodoxy, though remark that
one's reasons for maintaining such positions may have different levels of merit.}
Following that train of thought, the instantly-popular ChatGPT seemed a worthy candidate to test in the face of universal 
speculation 
over its capabilities and what it suggested in regards to AI's near-future trajectory.\footnote{The idea incubated until the end of April 2023, when a confluence of factors  opened the doors for this investigation. Serious, focused work began sometime in summer 2023, after the
groundwork conducted over the preceeding months 
exposed %
that this line of inquiry %
could feasibly contribute worthwhile insights.}
Note that, with that spirit and scope in mind, %
studying VQA-specific models or 
generative image networks would largely defeat the point. %

%% file: recognitionExperiments.tex
\label{sec:exps}

Our experiments revolve around the use of ASCII-art diagrams --- specifically, depictions
of boxes positioned on the page --- and GPT3.5's ability to identify and generate manipulations of them. It is worth a word 
on what inspired this choice, in particular the reasoning that leads this selection to be a compelling choice.

First, we realized that many different endeavors could utilize ASCII-art diagrams to illustrate concepts and designs. %
For instance, such drawings may be used as electrical-circuit diagrams, placement charts, 
and flowcharts in online help forums. Indeed, mini-languages like 
PIC \cite{DBLP:journals/spe/Kernighan82} and
Troff \cite{10.5555/107172.107183} have existed for decades to facilitate the creation of 
box-drawing, in addition to these depictions being easy to create by hand.
Thus, this is a form of ASCII-art for which GPT3.5 may have a substantial amount of
varied training data as compiled in the common-crawl dataset\footnote{\url{https://commoncrawl.org/big-picture/}}
and potentially other locations. Additionally, owing to its use in technical areas often as
an aid to facilitated understanding of a verbal description, these character-drawings may
have an appreciable amount of natural-language text describing their parts, which could
enable an LLM to connect aspects of the drawings to natural language descriptions.

Second, our interest in considering box-diagrams was ignited by preliminary
experiments we performed on OpenAI's ChatGPT web interface. Among our later trials in this 
phase, we asked ChatGPT to draw several towns, each with a school, place of worship, and houses, labeling
the parts according to our instructions.
We were intrigued by the model's ability to generate drawings so closely matching our specification, 
a feat not obviously performed by mere memorization of an example from the training set. During 
the interaction that followed, we requested ChatGPT draw roads between buildings in towns we specified,
and once again found the outcome to be noteworthy. 

Following from our preliminary investigations and later reflection on the format's reasonable merits, 
we have gone on to run a series of experiments that leverage randomly generated box-diagram depictions ---
featuring only boxes and no arrows --- to gauge ChatGPT's aptitude in tasks typically considered desirable
for vision related systems, such as the ability to cope with rotation, scale, 
``pixel'' noise, and translation of an image.  
If we find GPT3.5 can handle these tasks, then it is sufficient to say that the LLM is not \textit{entirely} incapable of ``doing well at ASCII-art''
in a sense, 
despite folk-knowledge suggesting otherwise (that any ability the model displays there is nothing more sophisticated
then rote memorization); there may be some ASCII-art where GPT3.5 fails to meet the bar for performance, but we 
are interested in there \textit{existing} a collection of reasonable illustrations --- some clearly visual structure ---
that it can handle in some fashion that is not  a raw look-up of exact examples from the training data. Even such mere
existence would be enough to add more nuance to the tale present in the 
general narrative.\footnote{Furthermore, it seems like quite a claim to indicate that any modern system that is
considered ``competent'' on visual tasks would also be confidently said to be capable across \textit{all} image 
families. That is, doing something well over a single or small family of images seems enough to deem that a 
system has reasonable visual capabilities. Setting the bar higher seems like it would invite in too many 
adversarial examples, even for a human's visual system.}
 
In the following subsection, we briefly detail the process we followed to
generate these pieces of ASCII-art, and in the subsequent subsections overview our experiments, their results,
and have a discussions so as to put outcomes into perspective. For the remainder of this document, 
when we discuss ``box-diagrams'', etc., this should be understood to be
\textit{ASCII-art depictions of} box-diagrams. Unless otherwise stated, we drop the verbal distinction between
``true-graphics'' and those rendered in character-art.

\subsection{Generation of ASCII-Art Box-Diagrams}
\label{subsec:exps.artGeneration}

In this subsection, we will overview how box-diagrams are generated under the default 
parameters. Various experiments consider varying parameters, details we will go over as 
they become relevant.

Our process starts with a blank 24-by-24 character canvas and progressively adds 
boxes to it. A box requires five pieces of information: a coordinate for its lower-left
vertex, a coordinate for its top-right vertex, and a name comprised of a single ASCII alphanumerical
character. Naturally, the position values should remain in the grid --- we do not consider illustrations
of boxes that are partially out of view. The naming we will comment on in a bit.

The actual process to determine the
box's coordinates involves two steps: a proposal phase and a rejection phase.

During the proposal phase, a start position and length are chosen for each axis independently, 
the former uniformly over the canvas, the latter via draw from a Poisson distribution with a parameter of
8. We found results with a Poisson at $\lambda=8$ to produce visually reasonable
illustrations with a desired variation in layout and (in a certain intuitive sense) complexity ---
for instance, results can range from well-aligned rows of roughly uniform boxes, to nested complexes arranged
in a scattered fashion. Adding a bit more sugar to the motivation, 
some back-of-the-envelope reasoning about the size of our canvas and the expected behavior of the 
parameterized Poison has some appeal in respect to roughly what we can expect to see; e.g., 
consider the mean and variance of the draws (both are $\lambda$) and what proportion of the canvas borders they make up
($\lambda/24$ is one-third in this case).
In order to avoid boxes which are too small to fit a name in --- or even be identifiable as a shape ---
if a length is proposed which is less than 3 (two characters for the boundaries and one for a space imbetween),
we draw again.

In the rejection phase, boxes that are too long to fit in the canvas from their starting point are 
thrown out. Additionally, boxes that would overlap others are rejected, as are those that
fill the corner of another box (which could occur when one box is nested inside another) since we reserve
those spaces as potential locations of name-labels.
When a proposal is tossed-out,
we generate another and try again. If after a thousand tries we are unable to find another acceptable
proposal, we simply give up and return the canvas as filled by boxes selected up to that point.\footnote{If we
succeed in adding a box after a certain number of attempts, the next proposal gets a fresh slate. It is not
1000 tries total across all attempts, but 1000 per attempt.} We continue to add boxes either until 
we can't find a working proposal after our maximum number of tries or when we successfully add 14 boxes to 
the canvas.

Largely, the process described up to this point is exactly how each box-diagram is generated.
A few additional details must be considered for trials that involve comparison of multiple
box-diagrams, however. 
While we are only concerned with investigating GPT3.5's visual abilities on some family of depiction, 
we want to perform reasonable diligence in ensuring that the behavior we observe is rooted in a 
reasonable notion of visual competency and not simply fueled by easy ways to cheat.
One precaution we take is to ensure that, for experiments that require comparing multiple 
box-diagrams to each other, each option to choose from has the same number of each type of character
present when noise is absent. This prevents occurrences like matching an option to a reference image
simply because the reference and choice both have the same number of characters, instead of paying 
attention to the structure of the depictions.\footnote{GPT3.5 has demonstrated poor counting ability
when asked to perform explicit arithmetic, but being able to handle numbers explicitly is different
than being effected by the presence of a different number of objects. A parrot might not be able to count
to 1000, but it can appreciate the difference between a nut that weighs one gram and one that weights
a kilogram when it comes time to fly off with them.} %
We generate depictions that meet these per-trial constraints using the following process:
To begin, we generate the first image with no limitations placed upon it. 
For the remaining illustrations to be generated, we keep track of the number and type of 
characters present on the canvas during the iterations of adding boxes, and clip the lengths 
to guarantee that the result would not produce greater than the allowed number of characters. If a canvas is produced that
has too few of any character, we reject it and try to produce another drawing, hoping for a better outcome; one
could say that we propose-then-reject on the canvas-level in addition to doing so on the box-level.

We draw the boundary of boxes using the dash-symbol (``-'') for length along the horizontal axis (x axis)
and the pipe-symbol (``|'') for length along the vertical axis (y axis). Early on, we placed a visually-aesthetic
plus-sign ("+") in the vertices of the boxes, but the character-matching constraints and the fact that 
any box has exactly four corners entails that doing so would require every drawing in an 
experiment to have the same number of boxes. Adding such a constraint ultimately struck us as unnecessary,
and perhaps detrimental, for what we were interested in investigating. %

In experiments, unless otherwise noted, we pad the right-margin of the ASCII-art with spaces so that each
line has the same number of characters.\footnote{Although, being some are white-space, not all of them are visible.} %
The alternative is to have
a ragged right-edge, which we speculate could make some tasks harder --- by introducing additional structure "noise" ---
and some tasks easier --- by providing a likely unique "key" on the right that could pick-out corresponding depictions
without having to examine the rest of the structure. Since the latter strikes us as a form of "cheating" and the former
appears like an added difficulty that would, as is appropriate, provide a more conservative bound on performance,
we opt for the uniform padding by default.

As to naming: as mentioned, we restrict names to being a single alphanumerical character. 
A box's name is drawn inside its boundary, at one of its four corners. The corner chosen is selected at random during the
generation process, and, again, boxes added later in the process are required not to overlap such positions. 
Within an experiment trial, if we show multiple diagrams (as we will in the recognition experiments),
we require that the same set of names appear in each illustration, a fact that also entails that the number of 
boxes present in each illustration is the same.
While the number is the same, the way we assign labels to the boxes is random for each diagram.
Unless otherwise noted, we refrain from including names in diagrams; we suspect this increases the difficulty
of the tasks, thus making any competence displayed by GPT3.5 more impressive and less questionable.

\subsection{Recognition Experiments on Diagram-Like Drawings}
\label{subsec:expRecogExp_diagrams}

Our first category of experiment concerned image recognition tasks. 
Each of the experiments conducted under this regime %
featured a prompt displaying a piece of reference ASCII-art, followed by a request to
GPT3.5 to specify which of three follow-up depictions match what the reference would look like if it were
to undergo some change; the options are presented in random order, so that the correct choice does not 
appear at any fixed letter.
Independent of the question posed to GPT3.5, the choices are generated
so that only one is based on the reference and the other two are freshly generated figures.
While one can imagine trials where more than one option is based on the reference but only 
one corresponds to the correct transform --- for example, identifying which is a quarter-turn clockwise
versus rotations of 180\textdegree and 270\textdegree --- we consider such trials to be a level of difficulty that is at least
imprudent to start with. Overall, we are interesting in identifying GPT3.5's ability to identify 
the same image pattern after it has undergone process typical for the vision data --- e.g., 
translation, enlargement, rotation, etc. 
If it is unable to succeed at such tasks despite only one option being derived from the reference art, 
then there seems reasonable to suppose having choices that are even more difficult to distinguish
from the reference would cause performance to degrade even further.

Taking the cue for Chain-of-Thought (CoT) Prompting \cite{wei2023chainofthought},
prior to posing the question with which we are primarily concerned to the model, we ask it a series
of warm-up questions to facilitate examination of the ASCII-art provided and consideration of
how the answer we eventually want could be determined. Two example prompts are shown in the appendix at
\cref{apdx:fig.diagramRecogPrompt1} and \cref{apdx:fig.diagramRecogPrompt2}.

Queries are issued to GPT3.5 once for each prompt, issued using OpenAI's API for \texttt{gpt-3.5-turbo} with no context other than our
prompt maintained between calls.
We draw responses with a temperature of zero, since
the solution space for most of the questions we are concerned with in this section have a comparatively 
small set of correct answers.\footnote{That is, if we asked the system to sample from a large, diverse set of answers,
intuitively we'd think a good chunk of that diversity would be outside a small collection of correct answers. If that happened,
its not obvious to us whether we should blame the system or blame our request. 
}
Despite the zero temperature, we have observed
in preliminary trials that responses were meaningfully 
diverse, with answers at times differing between them; sources of variance include OpenAI dispatching
requests to different running instances of the model, each instance responding with words that are somewhat different. %
We had initially begun experiments by querying three times per prompt and judging whether the
majority was correct. Ultimately we concluded that approach was not necessary for 
capturing the quantity we wanted to measure, the accuracy over the population. Furthermore, it
added complications to the analysis that would provide no obvious statistical benefits 
while increasing distractions and adding interpretation difficulties (e.g., is lack of performance to be considered
randomness per prompt or randomness per query).
While responses do vary when given the same query, taking many representative queries should 
reflect the population values we're concerned with for all the reasons typical to statistics.
This distinction between the response accuracy condition on one prompt and the 
response accuracy over the population will rear its head again when we discuss
analyzing human-drawn ASCII-art in a later section.

We found that responses we received to our prompt reliably had the answers we expected next to a 
corresponding sub-question numbering, for instance, 
\textit{"(1) The reference looks [...](2)[...](3) To determine which, I would[...](4) The answer is Choice A because [...]"};
strictly speaking, the replies did not comply fully to the shape we specified, since responses to part \textit{(4)} typically
included more than simply the name of the correct choice, but simple string parsing and regular expressions proved sufficient
to consistently extract the primary response of concern to \textit{(4)}. In cases where our extraction process ran into something
unexpected, we flag the circumstance and set it aside for review. Overall, we were able to rapidly extract GPT3.5's answers and
tally their correctness automatically, with only seven out of several thousand cases being marked for review.

In most cases, we do not indicate in the prompts what is contained in the ASCII-art, either in terms of the 
objects that are intended to be depicted (boxes) or the expected characters present.
For instance, in experiments involving translation,
we simply ask which one matches the target if it was shifted horizontally or vertically, not which one is, say, moved five spaces left in respect to the 
reference. 
Exceptions are as follows: %
\begin{enumerate}
\item{\textit{When the ASCII-art presented contains names.}
In this circumstance, while we keep the language largely vague, %
it may be possible that a careful reading implies 
the illustration likely contains certain varieties of shape; see the wording in \cref{apdx:fig.diagramRecogPrompt1}.}
\item{\textit{Experiments testing robustness to noise.}  Here, we explicitly refer to "boxes" existing in the reference ASCII-art (see the quoted language 
in \ref{subsec:expNoise}). This was done to help indicate what, in the ASCII-art, is not consider part of the
main pattern while providing minimal hints as to the syntax; that is, ideally the model would use this hint,
coupled with an "understanding" of what a box might look like, to determine where meaningful structure
is versus things it should ignore.
An alternative would have been for us to specify which characters may be noise and which
are not, however that degree of explicit syntax hint would likely subtract from how compelling the
experiment is.}
\item{\textit{Trials focusing on effects of rendering size.} We specify in the question whether the choices should be
scaled up or scaled down in respect to the reference, but other than that we do not indicate the content of the ASCII-art.
}
\end{enumerate}

Under this general setting, we examine whether GPT3.5 is capable of correctly identifying %
the ASCII-art corresponding to the reference after the following manipulations:

\subsubsection{Experiment Settings}
\label{subsubsec:expRecog_diagrams:expSettings}

\subsubsubsection{Matching Verbatim}
\label{subsec:expMatchVerbatim}

As a first gate to cross in our examination of GPT3.5's ``visual'' competence, we establish whether or not the network is capable of identifying a verbatim
copy of the ASCII-art. This entails, as the reader might guess, having among the choices one that is 
a duplicate of the reference, character for character.
For large sections of non-lingual input like ASCII-art\footnote{In the sense that ASCII-art is neither natural language or computer
code under reasonable guidelines.}, we'd like to verify that the network at least is capable
of the easiest form of recognition. 

Straight-forwardly enough, we request the model answer this:
\begin{quote}
Which choice has ASCII-art that matches the reference ASCII-art exactly?
\end{quote}

\subsubsubsection{Translation}
\label{subsec:expTranslation}

In order to determine the model's ability to match structures after translation,
we embed the ASCII-art we generate into a larger canvas and pick a random position for
the inner-canvas's bottom-left corner. Specifically, the larger canvas is taken to be
48-by-48 ( twice the length of the initial ASCII-art's canvas in each dimension) and the 
offset is chosen randomly in each dimension as an integer ranging from zero to 24 inclusive.
In order to ease interpretation of results, such as ensuring that all the cases we consider are
"interesting" at least in some minimal sense, we prevent the reference image and the correct choice
from having the same offset. In other words, we guarantee that the correct choice is some 
non-identity translation of the reference. We make no such restrictions on the other choices.

The question we posed to GPT3.5 for this tasks was as follows:
\begin{quote}
Which choice has ASCII-art that matches what the reference ASCII-art would look like after it has been moved left, right, up, or down? That is, which choice has ASCII-art that looks like the reference ASCII-art after a translation?
\end{quote}

\subsubsubsection{Rotation}
\label{subsec:expRotation}

We examine whether GPT3.5 is able to identify the illustration after a quarter-turn to the right (a 90\textdegree clockwise rotation). Our preliminary, informal trials suggested that this task is
generally difficult for the network, which is not entirely surprising since the location of characters
after the rotation change in a way that is relatively drastic compared to most (but not all)
text re-alignments. Driven by the desire to see if any glimmer of capability existed in this regard
(and not simply lazily give up because initial looks didn't strike us as stellar),
we investigated several different settings of ASCII-art side-length ($s$), maximum number of boxes ($B$) and Poisson-parameter ($\lambda$), as follows:
\begin{itemize}
\item{The default: $s=24$, $B=14$, and $\lambda=8$}
\item{Scaled roughly to a factor of $0.6$ the size: $s=15$, $B=9$, $\lambda=5$}
\item{Scaled roughly to a factor of $0.3$ the size: $s=8$, $B=5$, $\lambda=3$}
\end{itemize}
In combination with varying the size, etc., we investigate the impact of including the names in 
depictions; this means we examine six different parameter settings in total.
When the names are displayed, the corner of the box which they appear after rotation is, indeed, the
result after a quarter-turn right --- they are not put in the same corner as they appear in the reference (e.g., 
top-left, bottom-right, etc.) nor are they placed randomly.

Associated with this task was the following question:
\begin{quote}
Which choice has ASCII-art that matches what the reference ASCII-art would look like if we rotate the reference image 90 degrees clockwise? In other words, which choice shows what the ASCII-art would look like if it underwent a quarter-turn clockwise?
\end{quote}

\subsubsubsection{Noise} %
\label{subsec:expNoise}

It is not uncommon for images to have a certain amount of pixel noise ---  
small-scale, random alterations that are not obviously attributable to geometric transforms. 
Ideally, if GPT3.5 has non-zero aptitude for processing visual input,
it would be able to recognize depictions despite minor and semantically unimportant
variations. Pursuant to this goal of checking robustness to noise, we inject
random characters into the ASCII-art pieces --- both the reference and, separately, the choices\footnote{Elaborating
further, we do \textit{not} add noise to the reference image then paste it verbatim as a choice. We independently sample noise and
inject it into two separate copies of an initially noiseless canvas, one copy serving as the reference and the other listed among the options.} 
--- then
ask the model to identify which choice matches the reference.
We use a small set of ASCII special-characters as noise-elements,\footnote{Specifically the double-quotation sign, ``@'', ``*'', ``.'', and ",".} and only inject them in areas were a space was present in the original art.
We do not use characters that could be used as names or as boundaries of boxes, so given a character, we know immediately if it is noise we added.
We only replace white-space with noise in order to examine whether, under "nice" conditions, GPT3.5 is not 
thrown off (if it can't do this, trials with random insertion at any location would sensically be no better). %
More importantly,
this ensures that the structure identifying the reference ASCII-art is unarguably and unambiguously visible, avoiding any
concern that results might be poor because "the model could not have feasibly done well with so little information remaining", or otherwise
lose information critical to make a correct decision;
while certainly a far stretch from our present situation, one can think of this criterion as being analogous to the requirement that
adversarial injections for modern computer vision systems, which alter a network's predictions, do not to human eyes alter any of the image's meaningful 
content \cite{DBLP:conf/cvpr/EykholtEF0RXPKS18,khalid2021exploiting}.

We run experiments under two noise levels: $0.04$, which means that for each space character, there is a $4\%$ chance that it is 
converted to a noise character randomly chosen from the allowed character set, and $0.32$. We check that a piece of ASCII-art has at least
one noise character added, and repeat the injection process if not.\footnote{Strictly speaking, this increases the probability that a space is
replaced to be above the value set in our parameter. The effect should be minimal, however, and the number of times we have to resample due to this is
few if any.} In combination with this, we run experiments with either the default padding 
(i.e., guaranteed 24 characters per line) and maximum number of boxes (14), or with a ragged right-edge and at most six boxes. 
For the latter of these two, we are curious about performance when more variation is introduced %
(we remove the right-padding space after the last non-white-space character, whether it be from a box or from noise)
as well as there being "less signal" with fewer boxes present.

In our query to GPT3.5, we asked:
\begin{quote}
Ignoring the noisy characters injected into the depictions, which choice has ASCII-art which contains boxes that match the reference ASCII-art? That is, if we ignore characters that look like they are in the ASCII-artworks accidentally, which choice looks most like the reference ASCII-art?
\end{quote}

\subsubsubsection{Size/Scale}
\label{sec:expSize}

Like rotation and translation, a system capable of digesting visual input effectively 
should be able to recognize the same pattern when rendered at two different scales.
To study this, we generate ASCII-art at half its typical size then decide to display either the
reference or the choices, but not both, at double its initial size. %
We consider the choice of whether the reference is larger or the
choices are larger as a parameter. %
The ASCII-art initially generated has a 12-by-12 canvas, at most 7 boxes, and $\lambda$ of $4$;
when enlarged, the canvas is the standard 24-by-24 size.
In addition to the choice of which art to enlarge, we examine the impact of including names on the boxes,
in total making four different settings we run experiments under.

We ask the model one of two questions, choosing which of the two in accordance to our parameter settings:
\begin{quote}
Which choice has ASCII-art that matches what the reference ASCII-art would look like if we scaled the reference ASCII-art to double its size?
\end{quote}
or
\begin{quote}
Which choice has ASCII-art that matches what the reference ASCII-art would look like if we scaled the reference ASCII-art to half its size?
\end{quote}

\subsubsection{Results}

Our results for this section are listed in \cref{table:results.recognitionDiagrams}. 
In the table's parameters, "img. enlarged: ref." indicates the reference was shown at 24-by-24 scale and the options where 12-by-12; the
alternate, "img. enlarged; cho." does the reverse.

We list the observed accuracy on each of the trials we conducted, along with the number of samples
collected and a Clopper-Pearson confidence bound around the performance obtained.
Random guessing would have an expected performance of 33.3\%. 
Note that if one wants to convert the confidence bounds into a hypothesis tests in order to judge
whether the observed performance was consistent with random guessing (or worse), the proper significance-level for the one-side test would be 
$\alpha/2$; a hypothesis test with $\alpha=0.05$ would reject a null hypothesis of random performance \textit{more often}
than what one would conclude from our confidence bounds, and therefore one should consider our results more conservative.

To aid interpretation of results, we also show the performance of string edit-distance (Levenshtein distance).
In the ``unweighted'' results, the reference is compared to each choice and if the correct option is listed among those 
with minimum distance, we mark it as an instance of being correct. In the ``weighed'' case, the process is 
largely the same, except if there are $k$ ties for the choice with the minimum-distance ($k \in \{1,2,3\}$), 
a value of $k^{-1}$ is tallied for the accuracy, in contrast to the ``unweighted'' version which would simply add one.
While one could consider a more sophisticated version of edit-distance, such as one that weighs the cost of
replacement, removal, and addition by frequency of the characters,\footnote{Treating each trial as a transductive setting if not an 
inductive one. That is, for the sake of fair comparison, the frequencies may be based on the reference image and the choices as opposed to 
additional samples drawn from the population;
GPT3.5 was not granted access to additional samples for comparison.} adding that variety of cumbersome sophistication would eat-away at intuitive 
aspects of the edit-distance that partially motivate its use.
This baseline is more a sanity check than a head-to-head comparison with the LLM; even if there is some trivial
solution to a problem at hand, the question we're studying is whether GPT3.5 knows it. Edit distance maybe just
give us an idea of how much we should definitely weigh the possibility of GPT3.5 "cheating" on these tasks ---
a high-performance by edit-distance may further justify hedging endorsements of results from GPT.
This said, a counter-argument is that edit-distance could be considered a tool specific for the sort of job 
we're considering; it is questionable to compare something that has not been optimized for a particular problem type
(this is, \textit{possibly} GPT3.5) to something that has (edit distance).

On the note of comparison, we have not made any multi-test-like corrections
to \cref{table:results.recognitionDiagrams} to access the overall likelihood of a "false-positive" in the
sense of declaring some behavior to likely exclude behavior of pure random guessing.
Indeed, if continued to form groups of, say, 20 confidence intervals, each with exactly $\alpha=0.05$,
we'd expect one to fail to contain the parameter.
Here, we observe that with 13 confidence intervals with $\alpha \le 0.05$,
the probability that three or more fail to contain the parameter is less than 2.5\%;
this observation coupled with the fact that eight tests have confidence intervals
excluding accuracy of $\frac{1}{3}$ --- the performance if purely guessing ---
support the idea that not all the performance is random guessing and sample variability.\footnote{
Note that this can be caste in a Fisherian p-value-esque light as well as, separately, reinterpreted as a 
rule that over the long-term, over many trials, gives us a way to declare that at least one
 (or, adjusting thresholds, a few) of the bounds contain the parameter with a certain maximum error-rate.}

Considering the LLM's performance against the edit-distance, we see that there are cases where each performs better.
For scaling and noise trials, it is not surprising that the string distance does well, since for noise the reference and 
correct choice share all box-characters and letters at the same locations, and for scaling, judicious removal of rows and 
columns is sufficient for an exact match. While GPT3.5 seems to tackle those cases more effectively than random guessing,
it does not appear to be deploying as effective a strategy.
On the flip side, translation and, to a slight extent, rotation trials lean in favor of the network.
We speculate that translation was handled so well since tabbing and indentation of text is fairly common, 
and thus the model's training set likely has numerous examples covering it (in addition to white-space, in general,
playing an semantically unimportant role in most places it is employed). %
In experiments involving rotation, by raw-number GPT3.5 outperforms edit-distance, but the margins are 
minimal --- indeed, the difference between 35.2\% and 34.4\% accuracy is roughly 3 trials for a sample size of 395.
Observe
that the weighed edit-distance is lower than its unweighted alternative by 1 to 5 percent in those trials.

\begin{table*}[h!]
\begin{tabular}{ |c|c|c|c|c|c|c| }
\hline

\multirow{2}{*}{Experiment Type}  & \multirow{2}{3em}{Parameters}& \multicolumn{2}{|c|}{ GPT3.5 Acc. } &  \multicolumn{2}{|c|}{Edit Dist. Acc.} & \multirow{2}{3em}{Samp. Size} \\ \cline{3-6}
                                    &                              & Obs.   & CI , $\alpha=0.05$           & unweighted & weighed & \\ \hline \specialrule{1.0pt}{0pt}{0pt}
Verbatim                    &           ---                        & 99.8\% & [ 98.6\%, 100\%  ]  & 100\%  & 100\%& 400  \\ \specialrule{0.8pt}{0pt}{0pt}
\multirow{3}{3em}{Rotation} & size: 0.3                            & 34.0\% & [ 29.4\%, 38.9\% ] & 34.0\% & 29.9\%& 397 \\ \cline{2-7}
                            & size: 0.6                            & 35.2\% & [ 30.5\%, 40.1\% ] & 34.4\% & 32.3\%& 395 \\ \cline{2-7}
                            & size: 1.0                            & 34.5\% & [ 29.8\%, 39.4\% ] & 32.7\% & 31.6\%& 397 \\ \specialrule{0.8pt}{0pt}{0pt}
Translation                 &           ---                        & 90.5\% & [ 87.2\%, 93.2\% ] & 39.6\% & 39.0\%& 399 \\ \specialrule{0.8pt}{0pt}{0pt}
\multirow{4}{3em}{Scale}    & img. enlarged: ref., names shown: no & 39.6\% & [ 34.8\%, 44.7\% ] & 100\%  & 98.5\%& 396 \\ \cline{2-7}
                            & img. enlarged: ref., names shown: yes& 42.4\% & [ 37.5\%, 47.4\% ] & 100\%  & 100\% & 401 \\ \cline{2-7}
                            & img. enlarged: cho., names shown: no & 31.5\% & [ 27.0\%, 36.3\% ] & 100\%  & 98.5\%& 400 \\ \cline{2-7}
                            & img. enlarged: cho., names shown: yes& 38.0\% & [ 33.2\%, 43.0\% ] & 100\%  & 100\% & 400 \\ \specialrule{0.8pt}{0pt}{0pt}
\multirow{4}{3em}{Noise}    & noise lvl.: 0.04, padding kept: yes    & 44.0\% & [ 39.0\%, 49.0\% ] & 100\%  & 100\% & 398 \\ \cline{2-7}
                            & noise lvl.: 0.04, padding kept: no   & 42.1\% & [ 37.2\%, 47.1\% ] & 100\%  & 100\% & 399 \\ \cline{2-7}
                            & noise lvl.: 0.32, padding kept: yes    & 40.5\% & [ 35.6\%, 45.5\% ] & 100\%  & 100\% & 398 \\ \cline{2-7}
                            & noise lvl.: 0.32, padding kept: no   & 39.9\% & [ 35.0\%, 44.9\% ] & 100\%  & 100\% & 396 \\ \hline
\end{tabular}
\caption{\footnotesize Results from experiments determining the performance recognizing diagrammatic ASCII-art. 
    Following the experiment type and parameters, we list the observed accuracy and its corresponding Clopper-Pearson confidence bound at $\alpha=0.05$.
    To aid in building an intuition of results, we list the performance edit-distance achieves in recognizing the correct choice.
    In the parameters, "img. enlarged: ref." indicates the reference was shown at 24-by-24 scale and the options where 12-by-12; the
    alternate, "img. enlarged; cho." does the reverse. Recall that for the noise trials, when padding is removed, the maximum number of boxes we render is also decreased from 14 to 6. }
\label{table:results.recognitionDiagrams}
\end{table*}

\subsection{Recognition Experiments on Artistic Depictions of Animals and Machines}
\label{subsec:exps.recog.HumanDrawings}

\subsubsection{Motivation and Experiment Setup} %
\label{subsec:exps.recog.HumanDrawings.motivationAndSetup}

In the preceding section, \cref{subsec:exps.artGeneration}, we examined GPT3.5's prowess in 
vision with tasks that largely involved general biases and robustness useful for image-related tasks.
In concert with any aptitude in that regard, we would like to examine whether the model acquired any
pictorially-related semantic knowledge during its training. 

While the notion of GPT3.5 learning to identify the appearance of objects may seem
improbable, early checks of feasibility did not rule it out.  
In particular, using the ChatGPT's online %
interface, in at least several 
example runs, we found that parts of images were identified by the network.
For instance, in a small ASCII-art depiction of an owl, it correctly identified the 
beak. In early trials, we simply asked the network to include an ASCII-art arrow 
indicating the part of the figure where some object was located. Results varied from 
being reasonably accurate in our minds to less so. Motivating this section's experiment 
design was a mode of response that, while not strictly speaking incorrect, was ambiguous;
ChatGPT at times placed arrows that would, under reasonable interpretations,
identify a large swath of the depiction provided. 
For instance, when provided a small stick-figure and asked to indicate where the liver
would roughly be located, a large arrow was place below the figure aligning with the gap between the 
two legs, pointing (conceivably) to the whole body.

Motivated by our preliminary trials, we conducted ASCII-art recognition experiments 
that we deem not totally unreasonable lines of inquiry.
We manually selected 27 images from the ASCII-Art Archive (see \cref{subsec:relatedWorks.closest}),
covering several classes of animals --- dogs, cats, and birds --- as well as machines --- cars and planes.
The first two columns of \cref{table:recognitionExperiment.humanArt} indicate the type and number of each,
each row with a number in the second column corresponding to a distinct ASCII-art example we pulled out.

We set about constructing questions to pose to GPT3.5 regarding these depictions, opting to provide the ASCII-art, 
visually indicate a particular part of the art, then provide a multiple-choice question whose choices
are various feasible options. 
To reduce the ambiguity around visually identifying parts of drawings which we observed occurring with arrows
in our preliminary trials, %
we opted instead to copy the subset of characters in order to highlight them.
For instance, 
in \cref{figure:recognitionExp.humanDrawings.exampletoGPT}, 
we provide a depiction of a dog, followed later by the subset of characters
belonging to the back leg. Both depictions are padded with spaces to have the same number of character
per line, that is, there are precisely 14 character per line for both the illustration of the full
dog and its body-part; we perform this padding for every image we present to GPT3.5 in these
trials.

We will discuss the rest of what is depicted in 
\cref{figure:recognitionExp.humanDrawings.exampletoGPT}
momentarily, but first, in order for it to make sense, we need to discuss what the beginning
of our prompt to the LLM looks like.
As with our design in \cref{subsec:expRecogExp_diagrams}, 
we follow CoT guidelines and provide a series of questions leading up to the 
primary response we want. In addition to this, we inform
the system of what type of object is depicted in the full image (dog, cat, airplane, car, or bird),
and provide six examples of images, choices, and expected correct answers.
We provide this additional information since we deem this task to be potentially more challenging than that
of the prior section and --- in the case of naming the object shown --- subject to additional 
uncertainty in  respect to important details of what is meant to be depicted; we tried to select
ASCII-art that is relatively clear to human-eyes, but we recognize that, even for a person, a label may be 
needed to help clarify what is meant to be displayed. As for the examples:
\begin{enumerate}
\item{It is common practice to include 
such instances in prompts to GPT3.5, ex. \cite{wang2023selfconsistency,openAI_GPT_bestPractices}\footnote{It is %
often said, since the recent 
GPT-family models  can use few examples often to great effectiveness, that they are few-shot learners \cite{brown2020language}.}.}
\item{None of the images we went-on to query the network with were included with the exemplars.}
\item{Since the human-drawn ASCII-art tended to be far smaller than the multiple 24-by-24 images shown in most cases last section, we had more 
space available in GPT3.5's 4096-token context window to include such content.} %
\item{We did not rework the examples provided on a per-image basis nor on a body-parts (e.g., head, wing, wheel, etc.) basis. Instead, we kept 
them the same for all images, serving to demonstrate the general activity, in contrast to in-context examples otherwise 
found in the literature (\cite{liu2021makes}) which would, for instance, %
list exemplars of dog-and-body-part pairs
if the object we were querying about was a dog.}
\item{Finally, as can be seen in the example at \cref{figure:recognitionExp.humanDrawings.exampletoGPT}, we overview the inputs, choices for final answer, and
   correct responses for the multiple-choice question of primary interest. We do not illustrate potential CoT replies to the intermediate queries that are 
   posed to the network. }
\end{enumerate}

An example prompt for this task is in \cref{apdx:fig.humanASCIIArtPrompt1}. The exemplars --- not shown owing to limits of space (but we list
due to their relevance in our later discussion) ---
are the following in order, which we list in  the format ``object ; body-part ; number of choices to select from'' : (1) a stick-person ; 
the body ; 4 , (2) a car ; the wheels ; 2, (3) a cat; its tail; 4, (4) a bird; its head ; 3, (5) a dog ; its back leg(s); 4 (this is shown in
\cref{figure:recognitionExp.humanDrawings.exampletoGPT}), and (6) an airplane, its wings, 3.

In the questions we pose to GPT3.5, we provide exactly three choices in random order, only one of which is correct and the other two chosen randomly from a set of
feasible values.
All the choices, whether they be in the exemplars or the queries, are feasible
for the object in question. For instance, we do not show an instance
of a dog and provide ``wheels'' as one of the choices. %
The set of possible values are shown in the first columns of \cref{table:apdx.moreRecognitionResults.humanArt1} and \cref{table:apdx.moreRecognitionResults.humanArt2}
under headings indicating the object for which they might be used.
In the case of cars and airplanes (\cref{table:apdx.moreRecognitionResults.humanArt2}),
we include ``other'' as a choice, though never issue a question for which that is the proper answer;
naturally, we never inform GPT3.5 of this fact, so short of quite elaborate tracking on the server-side unbeknownst to us, %
all three choices are viable candidates as far as the model is aware.

\begin{figure}[h!]
\begin{verbatim}
[...]
\end{verbatim}
\begin{lstlisting}[basicstyle=\ttfamily\tiny,keepspaces=true,language={}]
Example 5:

EX_FULL_IMG:
```
  __    __
o-''))_____\\
"--__/ * * * )
c_c__/-c____/
```
OBJECT_IN_EX_FULL_IMG: a dog

EX_PART_IMG:
```

       c____/
```
EX_CHOICE_FOR_6:
    Choice A: front leg(s)
    Choice B: back leg(s)
    Choice C: tail
    Choice D: head
EXPECTED_ANSWER_TO_6_FOR_EX: Choice B
\end{lstlisting}
\begin{verbatim}
[...]
\end{verbatim}
\caption{\footnotesize One of the examples we provide as part of the prompt to GPT3.5 in the experiments
    of \cref{subsec:exps.recog.HumanDrawings}. The use of labels starting with "EX\_" are to help 
    reduce any chance of ambiguity as to the role the information plays in the prompt. The appearance of 
    the number six among the tags --- ``EX\_CHOICE\_FOR\_6'' and ``EXPECTED\_ANSWER\_TO\_6\_FOR\_EX'' --- are to
    indicate the sub-question of the prompt that, respectively, the choices and answer are for. See the example
    prompt at \cref{apdx:fig.humanASCIIArtPrompt1}. } %
\label{figure:recognitionExp.humanDrawings.exampletoGPT}
\end{figure}

\subsubsection{Results}

\small
\begin{table}[h!]
\begin{tabular}{ |c|c|c|c| }
\hline
\multicolumn{4}{|c|}{Per-Image Aggregated Results on Human Draw ASCII-Art}\\ \hline
Object & Img Num. & Stratified Avg. Acc. & Tot. Samp. Size \\ \hline
\multirow{5}{3em}{Birds}  & 1 & 59.2\% & 610 \\ \cline{2-4}
 & 2 & 53.8\% & 630 \\ \cline{2-4}
 & 3 & 53.1\% & 650 \\ \cline{2-4}
 & 4 & 26.5\% & 600 \\ \cline{2-4}
 & avg. \textbf{;} std & \multicolumn{1}{|c}{48.1\%} & \multicolumn{1}{@{}l|}{\textbf{;}\ \ \ 12.7\% } \\ \hline
\specialrule{0.8pt}{0pt}{0pt}
\multirow{5}{3em}{Cats}  & 1 & 39.9\% & 820 \\ \cline{2-4}
 & 2 & 37.2\% & 842 \\ \cline{2-4}
 & 3 & 42.6\% & 800 \\ \cline{2-4}
 & 4 & 33.2\% & 821 \\ \cline{2-4}
 & avg. \textbf{;} std & \multicolumn{1}{|c}{38.2\%} & \multicolumn{1}{@{}l|}{\textbf{;}\ \ \ 03.4\% } \\ \hline
\specialrule{0.8pt}{0pt}{0pt}
\multirow{6}{3em}{Dogs}  & 1 & 53.6\% & 890 \\ \cline{2-4}
 & 2 & 40.5\% & 840 \\ \cline{2-4}
 & 3 & 47.8\% & 860 \\ \cline{2-4}
 & 4 & 34.9\% & 810 \\ \cline{2-4}
 & 5 & 38.1\% & 788 \\ \cline{2-4}
 & avg. \textbf{;} std & \multicolumn{1}{|c}{43.0\%} & \multicolumn{1}{@{}l|}{\textbf{;}\ \ \ 06.8\% } \\ \hline
\specialrule{0.8pt}{0pt}{0pt}
\multirow{5}{3em}{Cars}  & 1 & 74.7\% & 380 \\ \cline{2-4}
 & 2 & 34.6\% & 450 \\ \cline{2-4}
 & 3 & 32.8\% & 400 \\ \cline{2-4}
 & 4 & 36.7\% & 420 \\ \cline{2-4}
 & avg. \textbf{;} std & \multicolumn{1}{|c}{44.7\%} & \multicolumn{1}{@{}l|}{\textbf{;}\ \ \ 17.4\% } \\ \hline
\specialrule{0.8pt}{0pt}{0pt}
\multirow{11}{3em}{Planes}  & 1 & 55.0\% & 430 \\ \cline{2-4}
 & 2 & 55.1\% & 430 \\ \cline{2-4}
 & 3 & 43.0\% & 420 \\ \cline{2-4}
 & 4 & 27.3\% & 430 \\ \cline{2-4}
 & 5 & 53.4\% & 400 \\ \cline{2-4}
 & 6 & 36.7\% & 400 \\ \cline{2-4}
 & 7 & 24.9\% & 370 \\ \cline{2-4}
 & 8 & 34.2\% & 400 \\ \cline{2-4}
 & 9 & 50.2\% & 420 \\ \cline{2-4}
 & 10 & 45.6\% & 390 \\ \cline{2-4}
 & avg. \textbf{;} std & \multicolumn{1}{|c}{42.5\%} & \multicolumn{1}{@{}l|}{\textbf{;}\ \ \ 10.7\% } \\ \hline
\end{tabular}
\caption{\footnotesize The STDs reported are from across the per-image proportions, not the per-image-part or per-query variability.}
\label{table:recognitionExperiment.humanArt}
\end{table}
\normalsize

Results from running our recognition experiments with human-drawn ASCII-art are shown in
\cref{table:recognitionExperiment.humanArt}.
Across sections, the image (ID) numbers refer to different ASCII-art; the row labeled ``1'' in the 
collection of rows labeled ``Birds'' is distinct from the similarly-numbered row associated with ``Planes''.
We report for each image the stratified accuracy and the total number of samples taken relating to that 
art-piece. While our sampling of performance over image parts was roughly uniform (i.e., we asked GPT3.5 to identify 
bird wings about the same number of times we asked it to spot bird legs), there are some non-negligible differences in the numbers 
collected for each, 
resulting from networking interruptions and other logistical factors.  
Tables \ref{table:apdx.moreRecognitionResults.humanArt1}
and \ref{table:apdx.moreRecognitionResults.humanArt2}
in the appendix list the precise counts.
While we did not anticipate performance noticeably correlating with the number of completed samples, %
for the sake of easing interpretation of our aggregated performance results, we simply took a uniform 
average over per-part performance in order to arrive at the per-object results shown in \cref{table:recognitionExperiment.humanArt};
we refer %
to this being a stratified accuracy, in order to contrast its uniform weighing of
performance on subsets to the alternative of dividing the cumulative number correct by the total, which would represent 
larger subsets more. 
The rows labeled ``avg. ; std'' report the average and standard deviation for the performance reported in the group of rows of which it is part.

In contrast to \cref{table:results.recognitionDiagrams} where each image was sent to GPT3.5 once, 
we post the same image multiple times, as shown in the rows of \cref{table:recognitionExperiment.humanArt}. 
This was done to make the best of the relatively small set of human-created ASCII-art available, each of which 
required additional effort to separate into parts. 
The net result of the sampling %
carried-out is that the variance for the estimate of each \textit{individual} image decreases.
However, trying to get an accurate
performance measure over the target distribution of ASCII-art more broadly is hampered at least by the relatively
small number of images considered.
To be more concrete, we can consider the following (rough) decomposition and approximation, where $\mathcal{X}$ is the distribution
over ASCII-art we'd like to judge performance on, $M$ is the model (GPT3.5), $l_{x}$ is the label of a sample $x$ (e.g., "wings", "legs", etc.),
and $S$ is our sample of images from $\mathcal{X}$:
\begin{align}
p_{x \sim \mathcal{X}}( M(x) = l_{x}) = \int_{\mathcal{X} } p(x)p( M(x) = l_{x} | x) dx \approx  \nonumber \\
\sum_{x \in \text{S}} \hat{p}(x)\hat{p}( M(x) = l_{x} | x) = |\text{S}|^{-1}\sum_{x \in \text{S}}\hat{p}( M(x) = l_{x} | x)  \label{eq1.l2}
\end{align}
By repeatedly querying the network about a single sample, $x$, we can achieve a better estimate of $p(M(x) = l_{x} | x)$, however
this is only part of the picture, and our estimate of $p_{x \sim \mathcal{X}}( M(x) = l_{x})$ is largely limited by other factors.
This all said, it is the right-hand side of %
 \cref{eq1.l2} that motivates us to report the average performances, despite its lack of robustness.\footnote{
With slightly more images to consider, we could bound the median between the extreme values, understanding that the
probability that all observed value are either above or below it is $2^{-n+1}$, for $n$ the sample size.
The surrounding context would need to be adjusted for the purposes of aiding our interpretation, however.}

This all said, it appears that at least some classes in the set considered provide
evidence (weakly) suggesting that detection 
performance is
better than at random.
Trying to ground ourselves to reality, we must try to rain on our own parade.
First, giving the small collection of ASCII-art images used, claiming that the performance
reflects an entire "class" as opposed to those specific images is a leap;
it may be more appropriate to say GPT3.5 knows something about \textit{those specific} images, 
as opposed to claiming it has acquired knowledge of properties inherent to each member in the
"class" which we're intuitively referring to.
Second, even if we admit performance is better than random, the mechanism used to achieve
such results are not necessarily impressive.
The image parts used likely correlate strongly with location in the ASCII-art ---
for instance, heads tend to be near the top, feet and wheels tend to be at the bottom, tails
near the left or right boundary, etc.
Memorization from the training set cannot be ruled out either, and as we alluded to when 
discussing the prior works, this is not a factor to entirely hand-wave away.

Quite importantly, and most likely a strong factor, 
is the influence of the examples we provided in the prompt to GPT3.5.
While we did not feature any of the ASCII-art that we went on to ask about,
we did have examples of heads, tails, back legs, and so forth. 
We did not curate the examples to each class --- all experiments received
the same demonstrations ---
but the point remains as to likely origins of the perhaps-present boost above 
random guessing that some results have a glimmer of. Future work could better investigate
isolating that variable and attempt to identify how it contributes.

Planes and cars deserve additional attention in respect to the influence of the exemplars, in particular
numbers (2) and (6). While it is not obvious how much it should be considered ``cheating'',
the location and general appearance of wheels in the car ASCII-art selected is not terribly dissimilar from the 
material in the prompt. More questionable is the performance of planes, since image 1---5 are
relatively similar to the exemplar, though each does have  
unique aspects; depictions at \cref{apdx:fig.humanASCIIArt.airplanes}.
Second, the trials included the choice ``other'', which, having not asked questions where that is a correct 
answer, only is represented in \cref{table:recognitionExperiment.humanArt} by how it reduces performance.
It is conceivable that an LLM--- having observed in the wild that open-ended and vague replies tend to be
more often %
wrong than right in multiple-choice tests --- rarely guessed ``other'', which would by itself put expected
performance in our tasks
near 50\% if one were to randomly guess each answer after removing that option.

All the qualifiers stated, we have \textit{not} shown that GPT3.5 simply fails 
on this recognition-\&-knowledge task. While a number of the measures are far from 
impressive, %
it is not the case that we have a slam-dunk for
the claims that, in contrast to its %
competence in pure-text tasks, GPT3.5 has (1) no aptitude whatsoever for purely visual activities,
and (2) no concepts whatsoever attach to ASCII-art and just %
matches it
verbatim from the training set %
without any knowledge of the composition other than its overall label. %

%% file: generationExperiments.tex
\label{sec:exp.gen}

Examining GPT3.5's abilities in respect to generation of ASCII-art, we
perform experiments requiring the model to transform reference images
in a particular fashion.

\subsection{ASCII-Art Used and Queries Issued}
\label{sec.exp.gen.queries}

In order to access the model's ASCII-art generation capabilities while 
anchoring the results to something we can access outcomes with, %
we follow a modification of the prompt-with-image-reference scheme detailed in
\cref{subsec:exps.artGeneration} and \cref{subsubsec:expRecog_diagrams:expSettings}.
We draw on the ASCII-art generation process from \cref{subsec:exps.artGeneration} to 
produce reference images that are shown to the network. Our prompts were once again 
shaped by Chain-of-Thought reasoning, providing warm-up questions leading to the 
ultimate request. Simultaneously, we tried to avoid revealing excessive details / step-by-step
instructions, in order to better gauge the degree to which GPT3.5 already had acquired a 
notion of what we were %
probing for. For some tasks --- such as translation---
we specified more details than others (e.g., no space above or on the left-margin) since
we found it to be the least cryptic and most pithy way to %
convey the outcome we desired.
The added specificity also helped to rule-out 
interpretations of our prompt that may have suggested alternate
translations (of which there are many) that could have otherwise been justifiably provided. 
This concern over mild ambiguity of intended outcomes ultimately was reflected in 
outputs we observed to \textit{a certain} degree for \textit{some} tasks, as we will overview in our qualitative assessment.

Before proceeding to overview results, we take a moment to detail the parameters used when 
generating the images for %
these experiments, and share the prompts ultimately
fed into the network. In contrast to most experiments conducted in \cref{subsec:expRecogExp_diagrams},
\textit{all} diagrams produced in our trials here featured name-labels. This was motivated by our
belief that (1) the generation task is inherently harder than the recognition task, and (2) providing
names to help anchor and minimally queue GPT3.5 as to structure would assist in the task, reducing 
chance of ``missing any interesting behavior'' that might otherwise be overlooked by setting up the LLM
for failure with tasks that are too difficult at the get-go. Certainly it would be a leap of logic to start
by demanding the LLM
generate Picasso's \textit{Guernica} in the style of Leonardo da Vinci,\footnote{Displaying in the style of Jackson Pollock might yield better outcomes.} then conclude the model lacks all ability if it fails.
Translation and verbatim-matching experiments were done with the same settings as before, in respect to the 
reference displayed, with exception of the naming already mentioned;
we requested that the model return the image without the extras spaces for the former, and
return the latter as-is.
For the size-trials, we displayed a half-size image and requested that the model scale it up by two. %
Noise-trials were conducted at the $0.04$ level with padding retained, and rotations were done at size $1.0$; see
\cref{table:results.recognitionDiagrams}.

In preliminary trials, we found that a fraction of the time GPT3.5 would reply to our prompt solely with
text or other non-ASCII-art content. In order to ease downstream analysis we intended to perform, we opted to 
add a light-weight mechanism for detecting such cases and re-issuing the query, hoping that the next reply
would result in the desired content being present. In particular, we reissued queries if either of the following held true:
\begin{enumerate}
\item{The response failed to be at least three lines long, the minimum height of our ASCII-art boxes.}
\item{There was no occurrence of the character ``|''  with the ``-'' character present either one space
up and over, or one space down and over. Recall that ``|'' and ``-'' are the vertical and horizontal
boundary markers for our boxes respectively, so this strategy essentially checked that there 
feasibly could be the depiction of a box-vertex in the content returned. In short, this 
provided a simple and fast method for detecting a feature indicative of pertinent ASCII-art being present.} 
\end{enumerate}
Queries were allowed to be reissued at most 14 times, after which the code simply gave up and marked the 
attempt as a failure. While the strategy outlined was a heuristic --- it not guaranteeing that entire, 
box-like depictions would be the final outcome --- situations where the outcome failed to contain 
ASCII-art germane to our purposes were few if any.\footnote{We don't recall any despite considerable 
manual review of outputs. We'd want to reinterrogate those materials, however, prior to stating that a 
failure (exhaustion of the 14-attempt budget) was \textit{never} reported.}

In the results to come, unless otherwise noted, we inspected 30 instances for each experiment.
We share illustrations that feature output ASCII-art which as extracted with a simple heuristic:
\begin{enumerate}
\item{Find and return the content in the last pair of triple-back-ticks ("```") present in the 
output. The network tended to output in this format, which if nothing else matched the 
style of presenting the input ASCII-art we used in our prompts (\cref{fig:exp.gen.queries.verbatimPrompt} 
and \cref{fig:exp.gen.queries.preambleForSeveralPrompts}, which we will overview momentarily).}
\item{If the first option fails to return something that appears to contain a box,
return all characters on
the line following the last line holding at least two consecutive alphanumeric characters. The detection 
of two letters or numbers next to each other is a loose heuristic to suss-out words. The logic of
taking everything following the last line-with-a-candidate-word was inspired by observations we 
made about the outputs during development-trials.
The vertex-finding heuristic described in the previous paragraph
was used to determine of whether the prior bullet's extraction attempts uncovered box-laden pictures. }
\end{enumerate}
We consider the lack of human effort in the extraction process to be both convenient and
reassuring, the latter as it reduces any concern around human biases impacting subtle
output characteristics like tabbing and existence of excess whitespace margins.

We close-out this 
subsection with copies of the prompts we presented to GPT3.5. 
See \cref{fig:exp.gen.queries.verbatimPrompt} for the 
prompt used in verbatim-generation trials, see \cref{fig:exp.gen.queries.translationPrompt} for
the translation-trial prompt, and see \cref{fig:exp.gen.queries.noisePrompt}, \cref{fig:exp.gen.queries.sizePrompt},
and \cref{fig:exp.gen.queries.rotationPrompt} for the noise, size, and rotation queries, respectively.
Figures \ref{fig:exp.gen.queries.translationPrompt} through \ref{fig:exp.gen.queries.rotationPrompt} make use
of the preamble text shown in \cref{fig:exp.gen.queries.preambleForSeveralPrompts}, while the content
in \cref{fig:exp.gen.queries.verbatimPrompt} does not, and is represented in its whole form modulo ASCII-art
specific to a query.

\begin{figure*}[h!]

\begin{subfigure}[h!]{\textwidth}
\begin{lstlisting}[basicstyle=\ttfamily\tiny,language={},escapeinside={(*}{*)},frame=single]
Instructions: I am about to show you a reference ASCII-art image. You are to return the ASCII-art image to me verbatim. Reference ASCII-art Image:
```
(*\normalsize \bfseries [...] *)
```
\end{lstlisting}
\caption{
\footnotesize Prompt used for trials concerned with generating verbatim copies of provided images. ASCII-art would be placed
where the bolded, bracketed ellipsis (\textbf{[...]}) are shown. }
\label{fig:exp.gen.queries.verbatimPrompt}
\normalsize
\end{subfigure}

\begin{subfigure}[h!]{\textwidth}
\begin{lstlisting}[basicstyle=\ttfamily\tiny,language={},escapeinside={(*}{*)},frame=single]
Instructions: I am about to show you a reference ASCII-art image, and then ask you questions about it and a task you must complete. The questions are numbered 1, 2, (*\it \bfseries and 3,*) and the task is indicated separately. The ASCII-art depicts a collection of boxes, some of which may be nested inside of other boxes. Note that in the ASCII-art, each box depicted is labeled with a unique name, which consists of an alphanumeric character and which appears in one of the box's corners.

Reference ASCII-art Image:
```
(*\normalsize\bfseries [...] *)
```
\end{lstlisting}
\caption{
\footnotesize Preamble text with overview of the tasks GPT3.5 is requested to complete, followed by the placement of where ASCII-art would be, as indicated by the bolded, bracketed ellipsis (\textbf{[...]}). The bolded, italicized text in the preamble is substituted with ``3 and 4,'' whenever the experiment involves four such questions.}
\label{fig:exp.gen.queries.preambleForSeveralPrompts}
\normalsize
\end{subfigure}

\begin{subfigure}[h!]{\textwidth}
\begin{lstlisting}[basicstyle=\ttfamily\tiny,language={},escapeinside={(*}{*)},frame=single]
(*\normalsize \bfseries [...Preamble from \cref{fig:exp.gen.queries.preambleForSeveralPrompts}...] *)

     Your job is to do the following, in order:
(1) Describe the reference ASCII-art image.
(2) What would you do in order to form a piece of ASCII-art that matches what the reference ASCII-art would look like if it had no blank areas at the top of it and no empty left margin? That is, how would you change the reference ASCII-art to look like it was translated so that there was not unneeded empty space around it (while preserving all internal spacing and structured)?
(3) What would the reference ASCII-art look like if it had no blank areas at the top of it and no empty left margin? That is, what would the reference ASCII-art look like after it has been translated so that there was not unneeded empty space around it?

Task: Provide ASCII-art that matches what the reference ASCII-art would look like if it was translated to have no blank areas at the top of it and no empty left margin. That is, show a modified version of the reference ASCII-art that has been translated so that there is no unneeded empty space around it (while preserving internal spacing and structure).
\end{lstlisting}
\caption{
\footnotesize Prompt used for trials of generating image translations.}
\label{fig:exp.gen.queries.translationPrompt}
\normalsize
\end{subfigure}

\begin{subfigure}[h!]{\textwidth}
\begin{lstlisting}[basicstyle=\ttfamily\tiny,language={},escapeinside={(*}{*)},frame=single]
(*\normalsize \bfseries [...Preamble from \cref{fig:exp.gen.queries.preambleForSeveralPrompts}...] *)

     Your job is to do the following, in order:
(1) Describe the reference ASCII-art image.
(2) In the reference ASCII-art, the only characters that should be present are "|", "-", alphanumeric characters, or whitespace. All other characters are noise that should not be present. List what characters are present in the reference ASCII-art that are noise.
(3) How would you remove noise from the reference ASCII-art so that only the characters that should be there are present?
(4) What would the ASCII-art look like if each character that is noise was replaced with a single space character?

Task: Provide what the reference ASCII-art would look like if you remove the noise and only leave the characters that should be present. Any single character you remove should be replace by a single space character.
\end{lstlisting}
\caption{
\footnotesize Prompt used for trials of generating de-noised versions of reference images.}
\label{fig:exp.gen.queries.noisePrompt}
\normalsize
\end{subfigure}

\begin{subfigure}[h!]{\textwidth}
\begin{lstlisting}[basicstyle=\ttfamily\tiny,language={},escapeinside={(*}{*)},frame=single]
(*\normalsize \bfseries [...Preamble from \cref{fig:exp.gen.queries.preambleForSeveralPrompts}...] *)

     Your job is to do the following, in order:
(1) Describe the reference ASCII-art image.
(2) What would you do in order to form a piece of ASCII-art that matches what the reference ASCII-art would look like if it was scaled up to double the size?
(3) What would the reference ASCII-art look like if it was enlarge by a factor of two? That is, what would the reference ASCII-art look like if it was made twice as large?

Task you must complete after answering the questions: Provide ASCII-art that matches what the reference ASCII-art would look like if we scaled the reference ASCII-art to double its size. That is, produce ASCII-art that has axis which are double the length of the reference, and which the images shown are enlarged respectively.
\end{lstlisting}
\caption{
\footnotesize Prompt used for trials of generating enlarged copies of images.}
\label{fig:exp.gen.queries.sizePrompt}
\normalsize
\end{subfigure}

\begin{subfigure}[h!]{\textwidth}
\begin{lstlisting}[basicstyle=\ttfamily\tiny,language={},escapeinside={(*}{*)},frame=single]
(*\normalsize \bfseries [...Preamble from \cref{fig:exp.gen.queries.preambleForSeveralPrompts}...] *)

     Your job is to do the following, in order:
(1) Describe the reference ASCII-art image.
(2) What would you do in order to form a piece of ASCII-art that matches what the reference ASCII-art would look like if it was rotated 90 degrees clockwise? That is, what you you do in order to depict the reference image after a quarter-turn clockwise?
(3) What would the reference ASCII-art look like if it was rotated 90 degrees clockwise? That is, what would the reference image look like after a quarter-turn clockwise?

Task: Provide ASCII-art that matches what the reference ASCII-art would look like if it was rotated 90 degrees clockwise. That is, show the reference ASCII-art after it has been rotated a quarter-turn clockwise.
\end{lstlisting}
\caption{
\footnotesize Prompt used for trials of generating image rotations.}
\label{fig:exp.gen.queries.rotationPrompt}
\normalsize
\end{subfigure}

\centering
\caption{Prompts used in the ASCII-art generation experiments of \cref{sec.exp.gen.queries}. Bolded, bracketed text of larger size indicates either places where a the preamble from \cref{fig:exp.gen.queries.preambleForSeveralPrompts} should be substituted in, or the place where ASCII-art for an instance of the query would be placed.}
\label{fig:exp.gen.queries.prompts}
\end{figure*}

\subsection{Qualitative Assessment of Results}
\label{sec.exp.gen.qualitativeAccessment}

In order to judge the performance of GPT3.5 at the generation tasks we set out in 
\cref{sec.exp.gen.queries}, we manually examined %
randomly generated queries 
for each of the transforms under analysis (e.g., see \cref{fig:exp.gen.queries.prompts}).
We investigated the use of systematic human trials (comparable to A/B testing, as we arranged them) %
and automated analysis (e.g., ROC curves of edit-distance differentiating real reference images 
versus fake given GPT3.5's output). Having investigated both, we believe that these are valuable
contributions for future work which we intend to carry-out, however they don't quite fully capture the 
grey-areas present in results we observed.\footnote{In our present treatment, we opt for the most effective
single avenue
for building an understanding, while the helpful lens that cannot stand on their own are left for
a %
lengthier manuscript.} 
In preliminary trials and human-backed A/B testing, we
observed that GPT3.5 did not simply fail or succeed at tasks, but appreciably often generated content 
along an orthogonal axis, where the outputs were not wrong per se, but also were not quite what we envisioned. %
Indeed, in light of early observations, we attempted to refine 
our prompt's verbiage to narrow the range of valid or near-valid interpretations
our requests could have --- resulting in language shown in \cref{fig:exp.gen.queries.prompts}. %
These improvements notwithstanding, clearly the potential for meaningful nuances --- as seen in earlier sanity-checks --- 
warrants an examination of outcomes
by a reasonably well-informed human 
to see whether curious behavior persists, and to what degree it
does if so. Here ``well-informed'' is meant in contrast to information-hiding we considered in ``blind trials''  
that were pursued to a certain extent, where knowledge of the query
issued was not \textit{explicitly}
revealed to the human reviewer (though how easily one could \textit{guess}
is another matter). %
In the remainder of this subsection, we examine the outcomes of our queries per transform, and
attempt to give a sense of successes, difficulties, and curiosities.
As with the recognition experiments, our focus will be on the final outcome, which here is the ASCII-art returned
by the network, and not the verbal responses provided in reply to our Chain-of-Thought prompting.

\subsubsection{Verbatim Trials}
\label{sec.exp.gen.qualitativeAccessment.verbatimTrials}

Examining the outcome of 31 trials for the verbatim experiments, 
by-and-large there is nothing to discuss but exact reproduction of the drawings we
provide.
Not only were the boxes generated precisely, but the right-padding spaces also 
were matched. An exception to that, which we observed only three times, was the right-padding
on the final line, which on those few occasions was not present, resulting in the line
ending where the bottom-right box did.
In roughly two of those cases, the top-line was also slightly shifted, lacking the full left-padding
needed to align with the boxes on the line below. %
The most interesting case we observed was that on one occasion, the network extended the 
boxes shown by one row; we show this in \cref{fig:.exp.gen.qualitativeAccessment.verbatimTrials.1}.

\begin{figure}
\begin{lstlisting}[basicstyle=\ttfamily\tiny,keepspaces=true,language={},escapeinside={(*}{*)},frame=single]
(*\bfseries The reference image:*)
        ---
    -- |  j|
   |  | ---     -----
   |  |        |     |
   |  |        |     |
   |4 |        |     |
    --         |     |
               |     |
 --------      |     |
|        |     |    U|
|        |  -   -----
|  -     | |n|
| |L|    | | |   ---  -
| | |    | | |  |  m||Z|
| | |    | | |  |   || |
|  -     | | |  |   || |
|C       | | |  |   || |
 --------  | |  |   || |
   ---      -    --- | |
  |   | --------      -
  |   ||        |    --
  |   ||        |   |  |
  |7  ||       W|   |z |
   ---  --------     -- 
(*\bfseries The image returned by GPT3.5:*)
        ---
    -- |  j|
   |  | ---     -----
   |  |        |     |
   |  |        |     |
   |4 |        |     |
    --         |     |
               |     |
 --------      |     |
|        |     |    U|
|        |  -   -----
|  -     | |n|
| |L|    | | |   ---  -
| | |    | | |  |  m||Z|
| | |    | | |  |   || |
| | |    | | |  |   || |
|  -     | | |  |   || |
|C       | | |  |   || |
 --------  | |  |   || |
   ---      -    --- | |
  |   | --------      -
  |   ||        |    --
  |   ||        |   |  |
  |7  ||       W|   |z |
   ---  --------     --
\end{lstlisting}
\caption{\footnotesize The example of interesting alterations to the the reference image
    produced by GPT3.5 during our verbatim generation trials. Notice the boxes C, L, n, m, and Z are
    one row longer in the results returned by the network than in the reference provided.}
\label{fig:.exp.gen.qualitativeAccessment.verbatimTrials.1}
\end{figure}

\subsubsection{Translation Trials}
\label{sec.exp.gen.qualitativeAccessment.translationTrials}

We studied the outcome of 30 trials that required the model to translate images.
On the whole, we consider results to be a mixed success, where cases of near perfection were present, 
as were a couple instances of irrelevant outputs, and, most often, indication that the model was clearly on the right track,
but not spot-on.
Of the 30 trials, only eight had cases where seemingly random code or prose was mixed in with the ASCII-art 
(or ASCII-art-like) content, and of those, only three featured images that did not have a clear semblance to
or signature of
the reference image.
Most commonly, images returned had excess whitespace on the periphery trimmed, as desired.
This apparent desired behavior was tempered by the occurrence of certain
``failure modes'', namely %
loss of boxes, failure to preserve inner-distances fully, 
or failure to fully preserve box boundary alignments (e.g., keep the "|" markers precisely one space above 
or below one another, as opposed to shifting one space left or right). 
This said, rarely did these factors combine to form character-soup, and remnants of the reference image remained
clearly visible, with a minimum of one to two boxes basically intact exactly;
in \cref{fig.exp.gen.qualitativeAccessment.translationTrials.1}, we provide an example of these 
middle-grade results that we regard as being closer to high-quality, and in 
\cref{fig.exp.gen.qualitativeAccessment.translationTrials.2} we 
give an instance that we deem closer to the spectrum's other end.
Notice that in both cases, the results have clear relation to the reference and 
do not have excess whitespace outside the illustrated area.
Further, observe that while certain boxes are apparently missing, there is not the occurrence
of fabricated boxes; even in cases (not shown) where random text appeared mixed-in with the ASCII-art, 
the text in question was clearly prose or code, not %
a pictorial invention that the model tried to pass-off as part of the reference image.
Finally, we note that three results were very close to exactly as desired, preserving the 
boxes almost exactly (a few box boundaries were mildly misaligned)
and performing close to the full translation desired (there may be one or two extra left-aligned spaces), while an additional two
retained the image structure, but failed to fully move the image left (i.e., excess left-padding
remained).
Overall we're given the impression that while the network is not spot-on
in its execution of this task, it possesses some non-trivial knowledge of %
what is entailed %
in the requested visual manipulation.

\begin{figure}[h!]
\begin{lstlisting}[basicstyle=\ttfamily\tiny,keepspaces=true,language={},escapeinside={(*}{*)},frame=single] %
(* \bfseries The reference image: *)

                              ------
                -            |     X|
               | | --------  |      |
               |A||w ---   | |      |
                - | |b  |  | |      |
                  |  ---   | |      |
                  |        | |      |
                   --------  |      |
                ------       |      |
               |t     |       ------
               |      |           --
               |      |          |R |
               |      |           --
                ------       ---   -
               --     ----- |   | |x|
              | s|   |  -- ||   | | |
              |  |   | |f |||  5| | |
              |  |   | |  || ---   -
              |  |   |W -- |
              |  |    -----
              |  |  --  -----  -----
              |  | |  ||     ||a    |
               --  |9 ||Y    ||     |
                    --  -----  -----

(* \bfseries Image returned by the model: *)
--------
|     X|
|      |
|      |
|      |
|      |
|      |
|      |
|      |
------
--     ----- |   | |x|
| s|   |  -- ||   | | |
|  |   | |f |||  5| | |
|  |   | |  || ---   -
|  |   |W -- |
|  |    -----
|  |  --  -----  -----
|  | |  ||     ||a    |
--  |9 ||Y    ||     |
   --  -----  -----
\end{lstlisting}
\caption{
\footnotesize An example of middle-grade result from the translation trials, this one leaning toward
the better-end of the quality spectrum. }
\label{fig.exp.gen.qualitativeAccessment.translationTrials.1}
\end{figure}

\begin{figure}[h!]
\begin{lstlisting}[basicstyle=\ttfamily\tiny,keepspaces=true,language={},escapeinside={(*}{*)},frame=single] %
(* \bfseries The reference image: *)

                                      ------
                       ------   ---- |     4|
                      |K     | |a   ||      |
                      |      |  ---- |      |
                      |      |       |      |
                      |      |        ------
                      |      |    ------
                      |      |   |B     |
                       ------     ------
                                      ------
                                     |      |
                                  -  |D     |
                                 |c|  ------
                      ---------- | |      --
                     | ------   || |     |  |
                     ||      |  || |     |  |
                     ||      |  || |     |  |
                     ||     L|  || |     | I|
                     | ------   || | ---  --
                     |V         | - |   |
                      ----------    |   |
                        -----     - |   |  -
                       |    i|   |0||  Y| |g|
                        -----     -  ---   -

(* \bfseries Image returned by the model: *)
 ------
------   ---- |     4|
|K     | |a   ||      |
|      |  ---- |      |
|      |       |      |
|------   ---- |      |
   ------
|      ||
|B     |
 ------

------
|D     |
c|      |
------------
 ------|  |
|      |  |
|      |  |
|     L|  |
| ------|
|V     |
------------
  -----  --
 |    i||0|
  -----

\end{lstlisting}
\caption{
\footnotesize An example of middle-grade result from the translation trials, this one leaning toward
the worse-end of the quality spectrum. }
\label{fig.exp.gen.qualitativeAccessment.translationTrials.2}
\end{figure}

\subsubsection{Noise Trials}
\label{sec.exp.gen.qualitativeAccessment.noiseTrials}

Over the 30 instances we manually examined for the noise-related generation experiment,
we found results tended to be reasonable but incomplete or mildly flawed.
In regard to reasonableness, unlike the ``squashing'' or loss of boxes that occurred %
across a number of the
instances from
\cref{sec.exp.gen.qualitativeAccessment.translationTrials}, the internal structure of the
images returned aligned with the reference images, save a minority of rows that maybe shifted 
one way or another. It is the latter shifting --- which most often goes toward the left
but on occasion toward the right\footnote{We suspect that the bias toward the left has roots in 
character-deletion examples or logic; in English and typical left-aligned displays for it,
the backspace button causes the line to shorten toward the
left. } ---  
which is primarily responsible for the ``mild flaws'' we referred to. Another source of 
mistakes was the removal of box names in addition to noise-characters: only 
one occasion had all names removed, but 20 instances had at least one name 
missing.\footnote{Our impression is that most often the majority of names are retained, however 
we need to revisit results to confirm that officially.} 
We observed that names were never invented. All names present in the output were present in the 
input and, for all except one box in one image, occurred in the same corner of the same box in the output
as the corresponding %
content in the reference picture.
This bias towards %
absence instead of %
inserting fake substitutes is consistent with behavior observed in 
\cref{sec.exp.gen.qualitativeAccessment.translationTrials}.

As to the perhaps most salient question of these trials, we observed the following in respect to the 
actual removal of noise characters:
We did not see any example where all noise characters were removed, though there was at least 
one case where the input was cleaned of %
a number of such marking, retaining only one of an original 16. 
Every instance we observed removed at least some of the undesired characters.
No case that we saw added more noise than was originally present, and moreover the strict subset that
remained were located in the same position in the result as the original, save a handful of cases where
individuals were shifted one space left or right with accordance to the row they were part of.
Finally, we did not (with our naked eyes) %
detect any noise character that tended to be removed
more than others, nor was it the case that GPT3.5 consistently failed to remove one type of character within an image.
That is, for a particular image, we saw cases where it removed \underline{some} instances of ``@'' and \underline{not others},
and moreover there was \underline{not} an obvious trend of removing all occurrences of ``*'' while retaining certain
appearances of ``@''; this statement holds true substituting ``@'' and ``*'' for other noise characters.
How undesired characters were treated did not obviously correlate with location in the image, though deeper analysis
 would be needed to rule-out anything but the strongest of relationships in that regard. 

Taken together, it is this tendency to decrease the noise present in an image without removing it 
totally that causes us to label the outcomes as ``incomplete''. Given the amount of structure that is retained in 
the images while the number of noise characters is reduced, however, we are not particularly disappointed in
the results --- we consider them a fair bit better than one would expect if GPT3.5 lacked any prowess in regard to 
handling such patterns. %
As before, though, one must have healthy hesitation before speculating
on the \textit{mechanisms} at play in achieving these results; de-noising may well be handled as primarily
a text-only venture, though the occurrences of non-trivial inter-line alignment of ``|'' characters may 
be an indication that it is is 
not treated solely as a 1-D, line-by-line task by GPT3.5. %

In \cref{fig.exp.gen.qualitativeAccessment.noiseTrials}, we show a middle-grade
result from our noise-trials, leaning toward the lower-quality end ever so slightly.
We choose this result as a representative since it is not far off from the middle of the 
quality spectrum, and it illustrates a number of the behaviors we observed.
Notice how the image returned by GPT3.5 has lines shifted right in addition to those
shifted left (the former is rarer across the data). In the result, the box that had 
been labeled ``Y'' in the input is lacking its name. The noise characters retained 
span the collection ``*'' and ``"'', but the reference image had more of both such
symbols. 
Observe further that in the output, the tightest bounding-box that contains all the 
remaining noise has not retained all the noise-characters originally present in that area; %
the line where ``5'' is located used to have an additional ``"'', and an occurrence of
``@'' has also been removed. That is, 
we do not see convincing evidence in this case of there being a relationship between a mark's position
and how it is treated. %
The details noted, overall the original reference structure is preserved and most 
noise characters are absent from the result presented by GPT3.5.

\begin{figure}[h!]
\begin{lstlisting}[basicstyle=\ttfamily\tiny,keepspaces=true,language={},escapeinside={(*}{*)},frame=single] %

(* \bfseries The reference image: *)
               --
(*\hl{,}*)        ---  |6 |
 -----  |(*\hl{,}*)  | | (*\hl{"}*)|    -
|T (*\hl{"}*)  | |   | |  |   | |
 -----  |Y  | |  |   | |
         ---  |  |   | |
     ------    -- (*\hl{"}*)  |E|
(*\hl{$\ast$}*)   |      |          -
    | ---- |           (*\hl{,}*)
  (*\hl{"}*) ||    ||  ---------
    ||  (*\hl{"}*) ||(*\hl{"}*)|j        |
 -- ||5   || |(*\hl{"}*)    (*\hl{$\ast$}*)   |
| h|| ---- | |         |
|  ||      | |    ---- |
|(*\hl{$\ast$}*) ||      | |   |   n||
|  ||     y| |   |    ||
|  | ------  |   | (*\hl{$\ast$}*)  ||
|  |         | (*\hl{@}*) |    ||
|  |         |   |    ||
 --  (*\hl{"}*)       |(*\hl{$\ast$}*)   ---- |
     -- (*\hl{.}*)     ---------
    |  | ---------    -
  (*\hl{"}*) | 0||k        |  |2|
  (*\hl{.}*)  --  ---------    -
(* \bfseries Image returned by the model: *)
               --
         ---  |6 |
 -----  |   | |   |    -
|T   | |   | |   |   | |
 -----  |   | |   |   | |
         ---  |   |   | |
     ------    --   |E|
    |      |          -
    | ---- |
     ||    ||  ---------
     ||    || |j        |
 --  ||5   || |     (*\hl{$\ast$}*)   |
| h  || ---- | |         |
|    ||      | |    ---- |
|(*\hl{$\ast$}*)   ||      | |   |   n||
|    ||     y| |   |    ||
|   | ------  |   | (*\hl{$\ast$}*)  ||
|   |         |   |    ||
 --  (*\hl{"}*)       |(*\hl{$\ast$}*)   ---- |
     --       ---------
    |  | ---------    -
     | 0||k        |  |2|
      --  ---------    -
\end{lstlisting}
\caption{A middle-grade result from our noise experiments. We added the highlighting of the 
     noise-characters present.}
\label{fig.exp.gen.qualitativeAccessment.noiseTrials}
\end{figure}

\subsubsection{Size Trials}
\label{sec.exp.gen.qualitativeAccessment.sizeTrials}

We found the thirty size-trials we scrutinized to be diverse and, of our experiments,
most subject to the moniker of ``not wrong per se, but not what I envisioned''.
Indeed, it was this collection of experiments that, in our early examinations and initial 
attempts at human blind-testing, lead us to more fully appreciate the modalities of 
pronounced, arguably-correct behavior that would be otherwise under-appreciated by
more narrowly focused analysis.

While our prompt (\cref{fig:exp.gen.queries.sizePrompt}) specified that the reference
image should be doubled in size, we rarely saw --- either in the rigorously reviewed
instances focused on here or in the preliminary glances --- instances where this \textit{exact}
result was achieved. We have been compelled by GPT3.5's performance at this task, however, 
in light of its consistent ability to enlarge images (if not consistently by a factor-of-two increase along \textit{both} axes)
or otherwise ``enlarge by doubling'' the picture. In the case of the first, where enlargement occurs, 
we are not terribly put-off by the fact that it is rarely an exact scaling of two; 
precise counting and arithmetic is appreciated as difficult for such LLMs, so we are 
reasonably content to settle for some enlargement, with greater positive impression as
the alteration approaches the desired scale. Reasonably often to our eyes, it seemed that
within an image, one axis generally grew while the other was kept the same size (``generally''
since a recollectable subset had boxes present which did not change size).\footnote{ In fairness to the %
model, we did accidentally use the singular form of ``axis'' in our prompt (see \cref{fig:exp.gen.queries.sizePrompt}), whereas
we meant the plural ``axes'' --- the rest of the query's language hopefully conveyed what 
we intended despite this oversight.}
The fact that this behavior occurred along either axis, sometimes vertically and sometimes
horizontally, is of some interest since we are under the impression that most text
GPT3.5 was trained on featured languages that are read horizontally;
we would need to revisit results and tabulate occurrences to confirm,
but our sense is that horizontal expansion was more frequent.
This one-axis tendency notwithstanding, there were certainly cases where boxes
grew in both extents. Furthermore, results appreciably often had a mix of boxes that
were enlarged and those that retained their original size. Reductions in size were rare.
This mix of behaviors across the instances (and at times within an individual image) 
leaves us uncomfortable commenting on the prevalence of each mode of operation, beyond noting that 
each display occurred appreciably often, and none %
seemed rare with the exception of shrinking.\footnote{To a certain extent, this rarity could be a result of the reference image
having been made smaller than is typical. Most commonly, input images contained at least one, and often
several, boxes that could not be made any smaller along one of their axes. This said, we suspect that, 
if one were to repeat attempts with larger inputs, the outcomes would still heavily favor enlargement,
or at least non-shrinking. }

A particular common pattern among these experiments was for box names to be repeated 
multiple time, either horizontally (seemingly most prominent), vertically, or, at times, in a rectangular patch within their
box. 19 cases exhibited this phenomena.
We mean ``repeated'' to indicate concurrent occurrences, %
without separating space, and as distinct from 
multiple separate, identically-named boxes being present. %
This name repetition tended to accompany some growth by the corresponding box.
To be clear, not all boxes that were scaled-up featured repeated names; some
retained exactly one letter as the label.
Speaking on an opposite phenomena, we counted 15 instances were the output lacked at least one name 
that was present among the input --- though not always lacking the corresponding
box (which would be displayed without its label). Of these, only seven were missing more than
one name, with an observable skew towards lower counts. %
Consistent with our observations in prior experiments, not once did we see a label
invented whole-cloth; all alphanumeric characters in the output were present somewhere among the input.

As hinted-at in the description up to this point, a common modality of growing the output 
was to repeat boxes that were present in the input, most frequently doing so in a fashion
that preserved elements of the original placement or structure (e.g., relative distances to 
other corresponding landmarks). For instance, (a) box(es) could be copied and translated 
straight down or straight across. In a couple instances, material tessellated, repeating
until reaching the maximum extent of the context window.\footnote{See the first observation in \cref{appendix:subsec:twoSmallObsAndConnections},
which connects here.}

Relatedly, we counted five instances of the 30 which feature the repetition of characters
until the context window was exhausted, displayed material either being boundaries of boxes that extended indefinitely
downward or a subset of the boxes tessellated until the end. Of these, all but one was missing
some box-label; that is, they contributed four to the aforementioned 15 where the outputs had
certain names absent.

Only a handful of times (approx. three) did the output seem largely divorced from the structure 
and naming of the input. Most often, of those two evils, the structure of the output was more visibly mutated than 
the naming and name-positioning. As already remarked, the letters used for names were often repeated 
multiple times --- arguably part of a valid interpretation of ``increasing the size'' of the reference image. 
The placement of names in the returned images roughly matched that of the input, in respect to the 
relative positioning of each (e.g., neighborhood relations were reflected); similar can be said of the boxes, 
though their size and exact positioning 
tended to be more visibly varied %
( although, that could be perceptual bias, it being easier to tell when larger components of an image are 
moved or altered in respect to other components, etc.).
En tot, extensive sharing of meaningful structural similarity between the input and out images struck us as the norm,
although we may be biased (either as experimenters specifically of human pattern-matchers generally) to 
underplay the differences. Setting aside delineation of whether results altered more or conserved more in 
respect to the input, certain substantive visually information was clearly retained in the typical case.

Briefly, we remark on departures present in the experiments %
discussed 
up to this point from patterns we noticed during our unofficial reviews
over the course of earlier development.
From preliminary analysis, a sub-trend we observed was doubling of boundary lines 
(e.g., two lines of ``-'' one on top of each other), but the 30 we final examined 
for this discussion --- using the language settle upon in \cref{fig:exp.gen.queries.sizePrompt},
which is refined from what we used in earlier examinations --- largely does not exhibit this. %
We at first felt mild frustration %
that patterns which were apparent %
 earlier %
are missing from the %
cases mulled over here, though perhaps that is a sign of improvement, the very reason we
refined the prompt. 
While we did modify the
language slightly in these trials compared to earlier attempts,
we are unsure whether the altered prompt lead to this outcome, or
that simply our memory is biased due to a 
combination of scouting more results previously, 
cherry-picking during sanity-checking, anecdotally recollection,
and the diversity of reasonable behavior that did not 
fully meet our intent (the many different flavors of 
``grey-area'' cases where the network 
is not entirely wrong per se).
Also yielded in older attempts were 
scattered examples of close-to-ideal behavior, however 
the aforementioned caveats apply just as well to those instances
as any other we garnered from then.

We provide three examples from our present size-trials but may add more in a future addendum, finding the
subtle but important 
diversity present in this experiment to be potentially worthy of more than three %
representatives.

As to example outcomes, see \cref{fig:exp.gen.qualitativeAccessment.sizeTrials.1},
\cref{fig:exp.gen.qualitativeAccessment.sizeTrials.2}, and \cref{fig:exp.gen.qualitativeAccessment.sizeTrials.3}.
\Cref{fig:exp.gen.qualitativeAccessment.sizeTrials.2} we consider a fairly good case, 
\cref{fig:exp.gen.qualitativeAccessment.sizeTrials.1} somewhere in the middle of the quality
range, and \cref{fig:exp.gen.qualitativeAccessment.sizeTrials.3} towards the lower end\footnote{Without stretching
on for thousands of characters, that is.} due to the greater degree it alters the arrangement of boxes in the output.
The first two figures show repeated letters, while the last does not, though it lacks the letter ``e'' from the reference.
\Cref{fig:exp.gen.qualitativeAccessment.sizeTrials.1} has repeated substructures
featuring the content labeled variations of ``Y'', ``T'', and ``l'';
the hints of a tessellation beginning in \cref{fig:exp.gen.qualitativeAccessment.sizeTrials.3}
may be reflected in the presence of boxes ``5'' and ``o'' labeled at the top and, seemingly copied and
translated, bottom of the output.
Crucially, notice that in all cases, the output features boxes that each are larger than the
corresponding entry from the input (although some tolerance may be required for 
\cref{fig:exp.gen.qualitativeAccessment.sizeTrials.2}, where
boundaries for ``00'' and ``gg'' are not fully
closed).

The actual shape of
\cref{fig:exp.gen.qualitativeAccessment.sizeTrials.2}
may be due to memorization --- supported mildly by the fact that suddenly letters are (roughly) centered in the
boxes instead of being in corners --- however the labels present, their positions, and
the mangle of boxes ``j'' and ``o'' in the results 
leads us to believe that this is highly improbable to have been taken verbatim 
from the training data.
The degree to which it is templated on prior examples is questionable, though after a certain threshold such
behavior is less suspect and more the bread-and-butter of (few-shot) machine learning.

\begin{figure}
\begin{lstlisting}[basicstyle=\ttfamily\tiny,keepspaces=true,language={},escapeinside={(*}{*)},frame=single]
(*\bfseries The reference image:*)

 -  --  ---
|w|| 3||  i|
 -  --  ---
 -  ------
| ||  -   |
| || | |  |
| || |l|  |
| ||T -   |
|Y| ------
 -
(*\bfseries The image returned by GPT3.5:*)

  --    --    ----
 |ww|| 33||  ii||  i|
 |ww|| 33||  ii||  i|
  --    --    ----
  --    --    ----
 |  ||  --    --   |
 |  || |  |  |  |  |
 |  || |ll|  |  |  |
 |  || |ll|  |  |  |
 |  ||TT--    --   |
 |YY||  --    --   |
 |YY|| |  |  |  |  |
  --    --    ----
  --    --    ----
 |  ||  --    --   |
 |  || |  |  |  |  |
 |  || |ll|  |  |  |
 |  || |ll|  |  |  |
 |  ||TT--    --   |
 |YY||  --    --   |
 |YY|| |  |  |  |  |
  --    --    ----

\end{lstlisting}
\caption{\footnotesize A representative input-output pair from our size-trials in ASCII-art generation.
The output has of course been trimmed to exclude the verbally responses to our Chain-of-Thought prompting
questions and focus on the ASCII-art output.}
\label{fig:exp.gen.qualitativeAccessment.sizeTrials.1}
\end{figure}

\begin{figure}
\begin{lstlisting}[basicstyle=\ttfamily\tiny,keepspaces=true,language={},escapeinside={(*}{*)},frame=single]
(*\bfseries The reference image:*)
        -
       |4|
  ---   -
 |M  | -  -
  --- |0||g|
       -  -
 --     --
|V |   |  |
 --    |  |
 ---   |  |
|G  |  |A |
 ---    --
(*\bfseries The image returned by GPT3.5:*)
            --
           |44|
      ------   --
     |MM  |  -  -
     |    | |00||gg|
      ------  -  -
      ------  -  -
     |VV  | |  |  |
     |    | |  |  |
      ------  |  |
      ------  |  |
     |GG  |  |AA|
     |    |  |  |
      ------   --
\end{lstlisting}
\caption{\footnotesize The second example from our size-trials for ASCII-art generation.}
\label{fig:exp.gen.qualitativeAccessment.sizeTrials.2}
\end{figure}

\begin{figure}
\begin{lstlisting}[basicstyle=\ttfamily\tiny,keepspaces=true,language={},escapeinside={(*}{*)},frame=single]
(*\bfseries The reference image:*)
 --       -
| 5|   - |o|
 --   |j| -
 -  -  -
|y||X|
 - | |    -
   | |   |e|
    -    | |
    --   | |
   |v |   -
   |  |
    --
(*\bfseries The image returned by GPT3.5:*)
 ----         ----
|  5 |     -- |  o  |
|    |    |  j  |     |
 ----     ----      ----
 ----     ----      ----
|  y  |  |  X  |
|     |  |     |
 ----     ----      ----
 ----     ----      ----
|  v  |  |     |
|     |  |     |
 ----     ----      ----
 ----     ----      ----
|  5  |  |  o  |
|     |  |     |
 ----     ----      ----
\end{lstlisting}
\caption{\footnotesize An third example from our size-trials for ASCII-art generation.}
\label{fig:exp.gen.qualitativeAccessment.sizeTrials.3}
\end{figure}

\subsubsection{Rotation Trials}
\label{sec.exp.gen.qualitativeAccessment.rotationTrials}

In our experiments with generating image rotations, we found
two failure modes to comprised virtually all instances, and 
the remaining handful not being more successful:
\begin{enumerate}
\item{repetition of boundary marks until the end of the context window,
    at times preceded by a few boxes %
    that appeared to be 
    copied from the reference image,}
\item{some shuffling of content, primarily names among box-shapes copied
    from the reference.}
\end{enumerate}
Of these two, the second had subcases where the result appeared to 
contain attempts at flipping the content along an axis, though the extent
this is true is unclear,
outside the most obvious cases of it among the 30 we scrutinized;
to a extent, we humans may be projecting structure onto 
random permutations. 
Beside attempts at relocating boxes, a fairly common case was that
names were moved, though the boxes present (shapes and positions)
where verbatim the same as the input; we counted 
seven
instances matching this description, eight if we tolerate a single box having 
its boundary mangled a bit.
Most common was case (1), %
11 instances 
exhibited this behavior, %
displaying a large quantity of 
repeated boundary markers --- either stacks of ``|'' seeming to be indefinitely 
repeating vertical boundaries\footnote{See the first observation in \cref{appendix:subsec:twoSmallObsAndConnections},
which connects here.} or regularly spaced ``-'' or ``|'' characters horizontally.

For cases outside the two noted before, the ASCII-art ultimately produced was clearly in a 
similar style as the input. Our temptation is to say that most often, in these remaining instances, 
the boxes present in the output correspond to the input in respect to number and shape, 
although they moved location; we refrain from asserting this, however, since we fear that we may be biased
toward seeing structure more prevalently than is well-supported --- so, %
without having yet endeavored to conduct a %
box-by-box, trial-by-trial
tally,\footnote{Life becomes even more complicated if we 
want to count boxes that are ``similar'' by some rubric, but not 
identical. For instance, boxes with the same name and position in 
respect to the input and output, but one being a single unit longer.} 
we leave a question mark around what proportion of the time this statement %
holds.

In the case of box naming, as with prior generation-experiments, names present in the output 
were present in the input, not fabrications by GPT3.5.
In an appreciable chunk of cases (perhaps a non-simple majority)
box-naming underwent some changes that, if we squinted, might look like a partially
successful flip, or two such flips along perpendicular axes; %
\cref{fig:exp.gen.qualitativeAccessment.rotationTrials.1} belongs to this category.
Trying to ground ourselves erring towards conservative performance estimation, more likely the names are a roughly random jumble:
for the number of boxes present, we suspect one could describe close to any random shuffle as some
``partially successful'' flip or a combination %
of two orthogonal ones. %
This all said and
though slim evidence to make a meal on, we came across at least three examples like
\cref{fig:exp.gen.qualitativeAccessment.rotationTrials.2}, whose movement of 
structures\footnote{The movement seen can either be characterized as swapping rows or --- as in 
other examples where the boxes were move visibly clipped near the boundaries --- wrapping-around of content
as is moves
down a fixed display window. } %
and naming alterations is consistent with an
approximately flip-like operation.

Names did not appear to be consistently
moved to destination-boxes whose distance from the image boundary was qualitatively similar to the origin-box's
boundary distance in the reference; 
specifically, names that were placed on boxes toward the center of the image sometimes were 
move to boxes touching the borders and vice versa. 
While perhaps inner-ness and outter-ness was conserved more often than it was not (or the opposite), 
the property certainly did not hold the same  in all circumstances. %
It is possible that some sort of preservation of qualitative
placement tends more often than not to occur --- we have neither totally ruled that out as a majority case, nor
have we %
gleamed that it is a safe bet.

Ultimately, we did not deem a single result of the 30 to be totally or largely a correct rotation.
This is not surprising: neither the poor performance observed in the recognition experiments for rotations
nor preliminary analysis we conducted during development provided fuel for optimism.   %
This all said, 
a comfortable majority of the time we observed that substantial visual substructures %
were preserved, and moreover
that the model made some attempt
to shuffle or alter the image while preserving its rough scale and origin (e.g.,
it did not appear that
our requests for rotations were confused for requests to perform a translation or resizing). %

\begin{figure}[h!]
\begin{lstlisting}[basicstyle=\ttfamily\tiny,keepspaces=true,language={},escapeinside={(*}{*)},frame=single]
(*\bfseries The reference image:*)
          -  ---     --
         |y||   |   |6 |
         | ||   |    --
         | ||f  |     -
          -  ---     | |
   ------            | |
  |      | --------- | |
  |      ||Y        || |
  |      | --------- | |
  |      | --------- | |
  |      ||   ---   ||G|
  |     c||  |   |  | -
   ------ |  |   |  |
 ------   |  |D  |  |
|      |  |   ---   | -
|      |  |        u||e|
|      |   --------- | |
|      |             | |
|      |             | |
|      | ---          -
|      ||   |
|O     ||   | -----   -
 ------ |  A||H    | |X|
         ---  -----   -
(*\bfseries The image returned by GPT3.5:*)
          -  ---     --
         |O||   |   |c |
         | ||   |    --
         | ||A  |     -
          -  ---     | |
   ------            | |
  |      | --------- | |
  |      ||H        || |
  |      | --------- | |
  |      | --------- | |
  |      ||   ---   ||e|
  |     X||  |   |  | -
   ------ |  |   |  |
 ------   |  |D  |  |
|      |  |   ---   | -
|      |  |        u||G|
|      |   --------- | |
|      |             | |
|      |             | |
|      | ---          -
|      ||   |
|y     ||   | -----   -
 ------ |  6||f    | |Y|
         ---  -----   -
\end{lstlisting}
\caption{
\footnotesize Example of our rotation trials where the result has the same box-structure as the 
the input, but the names largely appear flipped over the center horizontal axis.}
\label{fig:exp.gen.qualitativeAccessment.rotationTrials.1}
\end{figure}

\begin{figure}[h!]
\begin{lstlisting}[basicstyle=\ttfamily\tiny,keepspaces=true,language={},escapeinside={(*}{*)},frame=single]
(*\bfseries The reference image:*)
 --------      ---   --
|       E|    |  1| |  |
|        |    |   | |  |
|        |    |   | |  |
 --------     |   | | S|
               ---   --
           -------  ---
   -----  |       ||   |
  |     | |       ||   |
  |    Z| |  --   ||   |
   -----  | | u|  ||   |
    ----  |  --   ||   |
   |g   | |r      ||   |
   |    |  ------- |   |
   |    |          |   |
    ----           |x  |
 ---------          ---
| --     9| ------  ---
|| N|     || --   ||  K|
||  |     |||b |  | ---
||  |     |||  |  |  --
||  |     |||  |  | |Y |
| --      || --  d| |  |
 ---------  ------   --
(*\bfseries The image returned by GPT3.5:*)
 ---------  --  ---
| --     9||  ||  K|
|| N|     ||  || --
||  |     ||  |||Y |
||  |     ||  |||  |
||  |     ||  |||  |
| --      ||  ||  x|
 ---------  --  ---
    ----  -------
   |g   ||       |
   |    ||       |
   |    ||  --   |
    ---- | |r |  |
   ----- |  --   |
  |     | |u      |
  |    Z|  -------
   -----      ---
               --
               |S|
               |  |
               |  |
               ---
 --------      ---
|       E|    |  1|
|        |    |   |
|        |    |   |
 --------     |   |
               ---
\end{lstlisting}
\caption{
\footnotesize A second example from our rotation trials. Notice that the top and bottom row
of the results seem largely to have swapped compared to the input, and the center row also
underwent an apparent flipping of names. The reversal of nesting between the boxes ``r'' and ``u''
is, to our impression, an unusual occurrence, although this may be categorize as a specific
form of ``name jumbling'' which we appraised elsewhere.}
\label{fig:exp.gen.qualitativeAccessment.rotationTrials.2}
\end{figure}

%% file: conclusion.tex
Drawing inspiration from the prowess we expect a truly human-level intelligent agent to have
across multiple signal modalities,
in this work we examined GPT3.5's aptitude for visual tasks, where the inputs
feature content provided as ASCII-art.
In sharp contrast to the large majority of prior works, we made not attempt to
overtly distill the image content into a lingual summary.
We conducted experiments analyzing the model's performance on image recognition tasks after
various transforms typical
in visual settings, trials investigating knowledge of image parts, and image generation
tasks; in each of these categories of experiment, we found that while GPT3.5 had room for 
notable improvement, it was not totally lacking in regard to visual and pictorial aptitude.
Overall, while not passable for human-level performance, we were pleasantly surprised by
just how well GPT3.5 --- a model nominally trained on text-only input --- did on tasks 
designed to exercise visual and spatial capabilities.

%% file: futureWork.tex
A wide variety of future work can follow this effort, as made clear in the materials shared here
and from observations gleamed during our preliminary investigations.
Some remarks in this regard %
have already been made in the preceding document. Crumbs can also be found in the appendix.
At this time, we do not venture to include more in this release of the document
due to the limits of space (largely), time, and resources, as well as to deliver the 
information %
found in these pages to interested communities sooner rather than later.
We list briefly in this section, however, examples of the space open to travel; this is not an exhaustive
description.

Experiments could be extended, such as examining recognition tasks that feature
the shifting of box boundaries (or other boundary alterations),
repeating our existent generation experiments with broader modes of behavior,
\footnote{Ex: translating from the origin to somewhere else specific, rotation in more directions, etc.
We made forays into a number of these already, in preliminary and developmental stages.
Some appeared promising. For practical reasons alluded to
 here and earlier in our text, not everything that had the potential to add further insight was equally
feasible and/or %
resource-friendly
to explore at present with sufficiently satisfying rigor.}
or ramping-up choice difficulties (ex: by inducing greater commonalities).
Generation tasks featuring human-drawn ASCII-art can be carried out, 
following a similar pattern of part-extraction as done in our analysis of recognition.\footnote{
This setting would likely require less human effort per image that what engineering prompts for recognition needed,
which is an added convenience.}
Increasingly rich classes of automatically generated structure depicted in ASCII-art could be used alongside
our diagrams, each enabling new varieties of question.

New experiments can be added, like determining the correlation between the amount of noise
injected and confidence with which GPT3.5 declares something is or is not ASCII-art.\footnote{Confidence given as
percent or more qualitatively, like "more sure than unsure" or "very confident", etc., potentially 
using established scales from psychology or law.}
Ability to identify or induce certain spatial relations in images can be more rigorously explored.
Generating contrastive descriptions, ablation studies (of both images and prompts), imputation performance, 
and more extensive exploration of the dependent variable space are all in scope. %
Examination of how the network performs when manipulating art \textit{it} initially provides is worthy 
of certain attention as well.

Overall, numerous avenues of investigation exist down this path, and our earlier
efforts suggested that quite a few could expose intriguing nuances of behavior and 
mildly unexpected abilities possessed by GPT3.5 --- though some require more careful probing than others to see. %

%% file: appendix.tex
\section{Additional Remarks Regarding Related Work}

\subsection{Comment Regarding Approaches to Vision-Like Tasks Based on Symbol Manipulation}
\label{appendix:sec:symbolicVision}

This is a particularly difficult sentiment to back-up with a simple citation
since really it is a perspective informed by artisanal wisdom compiled in the field over ages. Furthermore,
it is more a matter of consensus and direction of the field than objective fact --- most computer vision
systems currently developed do not, in the traditional sense, fall into a symbolic category, but quality work
still is developed for spatial reasoning systems \cite{DBLP:conf/er/MitschRS11} and the
occasional effort for logic-driven vision does come forward \cite{DBLP:conf/ilp/DaiMWTZ17}.
The situation may be comparable to considering whether blimps will be the primary method of air-travel in the future:
at one point around the early $20^{th}$ century that perhaps may not have seemed unreasonable, but the author
speculates to guess most present-day aviation experts would doubt that that method of transport --- while having its place ---
will in the future dominate and displace alternatives like airplanes and helicopters.
Regarding the long-stretch trend, the older portion is reflected to a certain degree by
\cite{waltz1975understanding,DBLP:books/daglib/0001814} and the
newer to some degree by \cite{DBLP:journals/corr/abs-2210-01797}, for random samples along a gargantuan river;
these by themselves barely do justice, but we will have to content ourselves with them to spare further digressing
into historical matters of such immense scope.

\subsection{Two Small Observations/Connections Tied to Closely Related Works}
\label{appendix:subsec:twoSmallObsAndConnections}

In passing, two small connections we'd like to share are as follow:

For one, the authors of \cite{DBLP:journals/corr/abs-2306-07195} put their finger on 
the interesting drawing trend --- or trend in failing to draw --- we saw at \cite{internet_folk_wisdom_7}, where lines would continue in straight-forward, 
geometric ways. That is, there is a class of flawed attempts at drawing which have been observed that contain a clear structure, albeit a simple one. Ex: the attempt at drawing a person \href{https://web.archive.org/web/20230622062243/https://preview.redd.it/turns-out-chatgpt-is-really-good-at-ascii-art-v0-sp8t1i69mc5a1.png?width=659&format=png&auto=webp&v=enabled&s=6506911e5f478e53e73871e9d239581803f8450f}{\color{blue}found at this hyper-link}. 

Second, while not quite tying to our visually-concerned analysis, we take a
moment to give a shout-out to the survey \cite{Koco__2023}, which in their table 1 attempted
some estimate of whether ChatGPT was exposed
during training to elements of data they went on to query about. 

\newpage
\section{Additional Material Relating to Recognition Experiments}

\subsection{Example Prompts}

\begin{figure}[h!]
\tiny
\begin{lstlisting}[basicstyle=\ttfamily\tiny,keepspaces=true,language={}]
Instructions: I am about to show you a reference ASCII-art image, and then ask you a question about it in relation to three choices -labeled choice A, choice B , and choice C. Note that in each illustration, the objects depicted are labeled with a unique name, which consists of an alphanumeric character and which appears inside the object they label next to one of the object's boundaries. Your job is to do the following, in order:
(1) Describe the reference ASCII-art image.
(2) Describe each of the ASCII-art choices, A, B, and C.
(3) Describe how you would go about answering the question posed about the ASCII-art images to determine which choice is correct.
(4) Name which choice you believe is correct, only stating the name of the choice and nothing else.
Reference ASCII-art Image:
```
 -   -
|y| |9|
 -   -
 -
| |
| | ---
|l||  i|
 -  ---
```
Question: Which choice has ASCII-art that matches what the reference ASCII-art would look like if we rotate the reference image 90 degrees clockwise? In other words, which choice shows what the ASCII-art would look like if it underwent a quarter-turn clockwise?
Choice A:
```
 --  -
| i|| |
 -- | |
    |l|
     -
 --  -
|9 ||y|
 --  -
```

Choice B:
```
 -
|y|  -
 -  |i|
 --  -
| 9|
|  | --
|  ||l |
 --  --
```

Choice C:
```
 ---  -
|l  ||y|
 ---  -
 -
| |   -
| |  |9|
|i|   -
 -
```
\end{lstlisting}
\normalsize
\caption{\footnotesize Example prompt for the diagrammatic recognition tasks. This specific instance is a rotation trial with a size of roughly $0.3$.}
\label{apdx:fig.diagramRecogPrompt1}
\end{figure}

\newpage

\begin{figure}[h!]
\tiny
\begin{lstlisting}[basicstyle=\ttfamily\tiny,keepspaces=true,language={}]
Instructions: I am about to show you a reference ASCII-art image, and then ask you a question about it in relation to three choices -labeled choice A, choice B , and choice C.  Your job is to do the following, in order:
(1) Describe the reference ASCII-art image.
(2) Describe each of the ASCII-art choices, A, B, and C.
(3) Describe how you would go about answering the question posed about the ASCII-art images to determine which choice is correct.
(4) Name which choice you believe is correct, only stating the name of the choice and nothing else.

Reference ASCII-art Image:
```
 ---- .    ,   . ,"   -
|.@  | @"."   -----  | |
|" ..|..   ",|,    | |*|
|  " |*      |"".  | | |
|    | ---- ,|  @ "| | |
|  @ ||@.,*| |     | | |
 ---- |,,  | | **  | | |
 ---- |* @ |  -----   -
| -  ||". ,| ---- " ,
|| |.| ---- | * *| "  -
|| | | ." @.| ***|   | |
| -  |*     | .@,| @*|@|
|    | . .  |*., | * | |
|  ."|*     | @  |@, | |
 ----  .   .|    |    -
 .  @  .  ,  ----   ---
  * @  --------  *.|   |
   " ,|  @ , , |@  | . |
*  ,   --------  . |   |
   ----      ----- |,* |
" | @  |    | , @"||., |
 @|@@  | @ .|  @  ||   |
* |,  "|,". |   . ||.  |
   ----  *   -----  ---
```
Question: Ignoring the noisy characters injected into the depictions, which choice has ASCII-art which contains boxes that match the reference ASCII-art? That is, if we ignore characters that look like they are in the ASCII-artworks accidentally, which choice looks most like the reference ASCII-art?

Choice A:
```
  --       ,  *  ------
 | .| ------- " |    @,|
 |  ||    @  |  |      |
"|  || .  . .|""|  --- |
 |.*||  @"   |  | |," ||
 |@ ||" .    | @| |   ||
 |  ||   . **| "|  --- |
* --  -------   |@@ @  |
  " *    @ * ,   ------
 @, ----       ---- *,
  ,| @ @|    ,|    | --
   |",  |"" . |  *@||,*|
 " |  . |,    |*  .|| ,|
"  |    |"    |   .||  |
,. | *  |" @  |   .||  |
   |@   |"    |@.  || "|
, @ ----   .   ---- | ,|
     "*  ------- @  |. |
  * * "@| *  ", |   | .|
*     " |".   . | - |. |
    ,   |  ---- ||*|| @|
 ----   |"|,"  |||,| --
| ,  | "|  ---- ||*|,,.
 ---- ,, -------  -   ..
```
\end{lstlisting}
\normalsize
\begin{verbatim}
[...]
\end{verbatim}
\caption{\footnotesize Another example for diagrammatic recognition, this time for a noise-trial with names not shown and a noise level of $0.32$. Owing to limits of space, we only show one of the three choices presented.}
\label{apdx:fig.diagramRecogPrompt2}
\end{figure}

\newpage

\begin{figure}[h!]
\tiny
\begin{lstlisting}[basicstyle=\ttfamily\tiny,keepspaces=true,language={}]
I am about to show you two pieces of ASCII-art then ask you as series of questions about them.

The first piece of ASCII-art is a full image, and the second is part of that image which has only some of its non-whitespace characters retained while the rest have been blanked out. The full image will be labeled FULL_IMAGE above it, and the mostly blanked-out image will be labeled IMAGE_PART. The ASCII-art will be labeled to indicate which is which. In addition to these pictures, we will provide the name of the object that the ASCII-art in FULL_IMAGE is meant to depict, providing it immediately following the tag OBJECT_IN_FULL_IMAGE.  The tag OBJECT_IN_FULL_IMAGE and the name of the object follows the full image but preceeds the other ASCII-art.

FULL_IMAGE
```
             ______--------___
            /|             / |
 o___________|_\__________/__|
]|___     |  |=    ||  =|___ |"
//   \\    |  |____||_///   \\|"
|  X  |\--------------/|  X  |\"
 \___/                  \___/

```
OBJECT_IN_FULL_IMAGE a car

IMAGE_PART
```

  ___                    ___
 /   \                  /   \
|  X  |                |  X  |
 \___/                  \___/

```

Please answer the following questions, numbered one through six, in order:

(1) Describe the ASCII-art shown in FULL_IMAGE, indicating the shape of its parts and what they are comprised of.

(2) Describe how you would expect an ASCII-art depiction of the type of thing indicated by OBJECT_IN_FULL_IMAGE to look like. Indicate its shape and what parts you expect to be present.

(3) Describe the ASCII-art shown in IMAGE_PART, indicating the shape of its parts and what they are comprised of.

(4) For each of the following sub-parts --- 4.1, 4.2, and 4.3 respectively --- describe what characters in FULL_IMAGE you believe represent them, if any:
    (4.1) wheel(s)
    (4.2) other
    (4.3) body

(5) Describe how you would determine which part of FULL_IMAGE the art in IMAGE_PART corresponds to.

(6) Of the following three choices --- Choice A, Choice B, or Choice C --- which provides the best name for the part of FULL_IMAGE that is shown in IMAGE_PART ?
    Choice A: wheel(s)
    Choice B: other
    Choice C: body

EXAMPLES:
The remainder of this prompt has examples of full images (labeled EX_FULL_IMG), parts (labeled EX_PART_IMG) and names of objects shown in EX_FULL_IMG (labeled OBJECT_IN_EX_FULL_IMG), followed by the tag EX_CHOICE_FOR_6 listing choices provided to choose a name for the image in EX_PART_IMG and then the tag EX_EXPECTED_ANSWER_TO_6 indicating the letter of the correct choice shown among those in EX_CHOICE_FOR_6.
\end{lstlisting}
\normalsize
\begin{verbatim}
[...]
\end{verbatim}
\caption{
\footnotesize An example of a prompt we provide for the trials judging performance on human-produced ASCII-art of animals and machines.
To keep reasonably within the space of a page, we do not show the six exemplars that were included with the text, instead placing ``[...]'' in the position
where they would be located. Per-prompt, the order of choices is randomly selected, but within a prompt, we keep the ordering under part-4 the same as they are listed in part-6.}
\label{apdx:fig.humanASCIIArtPrompt1}
\end{figure}

\subsection{Additional Tables from Recognition Experiments on Depictions of Animals and Machines}
\label{apdx:expRecog.drawnArt.moreTables}

In this section, we provide additional tables to complement \cref{table:recognitionExperiment.humanArt}.
\Cref{table:apdx.moreRecognitionResults.humanArt1} features information about trials on animal ASCII-art, 
while \cref{table:apdx.moreRecognitionResults.humanArt2} does so for machines. Each table is split into 
rows relating to a particular type of object  as indicated by a reading. The columns, running from left to right, are:
\begin{itemize}
\item{The name of the part the group of rows pertains. This is what the part displayed truly was, and thus was the 
     thing GPT3.5 had to guess in order to be correct.} 
\item{The image (ID) number the specific row refers to. E.g., the first non-header row of \cref{table:apdx.moreRecognitionResults.humanArt1}
    relates to the performance on the head of bird-image number 1.}
\item{The observed accuracy for when the true object and body part are as indicated.}
\item{An $\alpha=0.05$ Clopper-Pearson bound on the accuracy}
\item{And finally, the sample size.}
\end{itemize}
The rows that list ``avg.; std'' are followed by the average and standard deviation of the performance across images for the group it is a part of, respectively.
Within one object class heading, the same image ID number refers to the same (complete) image, albeit different 
secondary image (the one that highlights the body part in question) across different rows; for instance, the row labeled 
``Head'' with ID ``1'' under ``Birds'' relates to the same complete image (but different highlighted subset of 
characters) as the row called "Wing(s)", ID ``1'' under that same heading.

As discussed in \cref{subsec:exps.recog.HumanDrawings.motivationAndSetup}, the machines (\cref{table:apdx.moreRecognitionResults.humanArt2})
had three options listed in their prompts to GPT3.5: the two parts listed in the first column of the table, and ``other'', with the order of presentation randomized.

Important to note when interpreting these tables --- and in stark contrast to those shown in \cref{table:recognitionExperiment.humanArt} ---
it, hypothetically, would be possible for the LLM to always guess the same answer and have \textit{some} row of these tables listed as
achieving 100\% performance. For instance, in anytime the ``back leg(s)'' option was available, GPT3.5 guessed the corresponding letter, 
the corresponding row-groups in \cref{table:apdx.moreRecognitionResults.humanArt1} would have 100\% listed for each image, since whenever the
part indeed depicts that, the answer is not wrong. The cost of this approach would be displayed in the other part-classes, which
would of course have a higher rate of being wrong as a result of being ignored in such a large case. 
The aggregated performance on each image across its parts is shown in \cref{table:recognitionExperiment.humanArt}.

\newpage

\tiny
\begin{table}[h!]
\begin{tabular}{ |c|c|c|c|c| }
 \hline
\multicolumn{5}{|c|}{Birds}\\ \hline
Part & Img Num. & Acc. & CI, $\alpha=0.05$ & Samp. Size \\ \hline
\multirow{5}{3em}{Head}  & 1 & 57.1\% & [ 50.2\%, 63.9\% ] & 210 \\ \cline{2-5}
 & 2 & 98.1\% & [ 95.2\%, 99.5\% ] & 210 \\ \cline{2-5}
 & 3 & 50.0\% & [ 43.2\%, 56.8\% ] & 220 \\ \cline{2-5}
 & 4 & 26.4\% & [ 20.7\%, 32.7\% ] & 220 \\ \cline{2-5}
 & avg. \textbf{;} std & \multicolumn{1}{|c}{ 57.9\% } & \multicolumn{2}{@{}l|}{\textbf{;}\ \ \ 25.9\% } \\
\specialrule{0.8pt}{0pt}{0pt}
\multirow{5}{3em}{Leg(s)}  & 1 & 66.5\% & [ 59.5\%, 73.0\% ] & 200 \\ \cline{2-5}
 & 2 & 11.0\% & [ 07.0\%, 16.2\% ] & 200 \\ \cline{2-5}
 & 3 & 36.8\% & [ 30.4\%, 43.6\% ] & 220 \\ \cline{2-5}
 & 4 & 33.0\% & [ 26.5\%, 40.0\% ] & 200 \\ \cline{2-5}
 & avg. \textbf{;} std & \multicolumn{1}{|c}{ 36.8\% } & \multicolumn{2}{@{}l|}{\textbf{;}\ \ \ 19.8\% } \\
\specialrule{0.8pt}{0pt}{0pt}
\multirow{5}{3em}{Wing(s)}  & 1 & 54.0\% & [ 46.8\%, 61.1\% ] & 200 \\ \cline{2-5}
 & 2 & 52.3\% & [ 45.5\%, 59.0\% ] & 220 \\ \cline{2-5}
 & 3 & 72.4\% & [ 65.8\%, 78.3\% ] & 210 \\ \cline{2-5}
 & 4 & 20.0\% & [ 14.4\%, 26.6\% ] & 180 \\ \cline{2-5}
 & avg. \textbf{;} std & \multicolumn{1}{|c}{ 49.7\% } & \multicolumn{2}{@{}l|}{\textbf{;}\ \ \ 18.9\% } \\
\specialrule{0.8pt}{0pt}{0pt}
\specialrule{0.4pt}{0pt}{0pt}
\multicolumn{5}{|c|}{Cats}\\ \hline
Part & Img Num. & Acc. & CI, $\alpha=0.05$ & Samp. Size \\ \hline
\multirow{5}{3em}{Back Leg(s)}  & 1 & 24.5\% & [ 18.7\%, 31.1\% ] & 200 \\ \cline{2-5}
 & 2 & 42.0\% & [ 35.1\%, 49.0\% ] & 205 \\ \cline{2-5}
 & 3 & 25.5\% & [ 19.6\%, 32.1\% ] & 200 \\ \cline{2-5}
 & 4 & 32.0\% & [ 25.6\%, 38.9\% ] & 200 \\ \cline{2-5}
 & avg. \textbf{;} std & \multicolumn{1}{|c}{ 31.0\% } & \multicolumn{2}{@{}l|}{\textbf{;}\ \ \ 07.0\% } \\
\specialrule{0.8pt}{0pt}{0pt}
\multirow{5}{3em}{Front Leg(s)}  & 1 & 29.5\% & [ 23.1\%, 36.5\% ] & 190 \\ \cline{2-5}
 & 2 & 20.9\% & [ 15.7\%, 26.9\% ] & 220 \\ \cline{2-5}
 & 3 & 32.3\% & [ 26.1\%, 38.9\% ] & 220 \\ \cline{2-5}
 & 4 & 26.8\% & [ 20.7\%, 33.7\% ] & 190 \\ \cline{2-5}
 & avg. \textbf{;} std & \multicolumn{1}{|c}{ 27.4\% } & \multicolumn{2}{@{}l|}{\textbf{;}\ \ \ 04.2\% } \\
\specialrule{0.8pt}{0pt}{0pt}
\multirow{5}{3em}{Head}  & 1 & 65.9\% & [ 59.2\%, 72.1\% ] & 220 \\ \cline{2-5}
 & 2 & 57.5\% & [ 50.4\%, 64.3\% ] & 207 \\ \cline{2-5}
 & 3 & 69.4\% & [ 62.2\%, 76.1\% ] & 180 \\ \cline{2-5}
 & 4 & 55.9\% & [ 48.9\%, 62.7\% ] & 211 \\ \cline{2-5}
 & avg. \textbf{;} std & \multicolumn{1}{|c}{ 62.2\% } & \multicolumn{2}{@{}l|}{\textbf{;}\ \ \ 05.7\% } \\
\specialrule{0.8pt}{0pt}{0pt}
\multirow{5}{3em}{Tail}  & 1 & 39.5\% & [ 32.9\%, 46.5\% ] & 210 \\ \cline{2-5}
 & 2 & 28.6\% & [ 22.6\%, 35.2\% ] & 210 \\ \cline{2-5}
 & 3 & 43.0\% & [ 36.0\%, 50.2\% ] & 200 \\ \cline{2-5}
 & 4 & 18.2\% & [ 13.3\%, 23.9\% ] & 220 \\ \cline{2-5}
 & avg. \textbf{;} std & \multicolumn{1}{|c}{ 32.3\% } & \multicolumn{2}{@{}l|}{\textbf{;}\ \ \ 09.7\% } \\
\specialrule{0.8pt}{0pt}{0pt}
\specialrule{0.4pt}{0pt}{0pt}
\multicolumn{5}{|c|}{Dogs}\\ \hline
Part & Img Num. & Acc. & CI, $\alpha=0.05$ & Samp. Size \\ \hline
\multirow{6}{3em}{Back Leg(s)}  & 1 & 50.9\% & [ 44.2\%, 57.5\% ] & 230 \\ \cline{2-5}
 & 2 & 40.0\% & [ 33.3\%, 47.0\% ] & 210 \\ \cline{2-5}
 & 3 & 60.9\% & [ 54.2\%, 67.2\% ] & 230 \\ \cline{2-5}
 & 4 & 41.0\% & [ 34.1\%, 48.2\% ] & 200 \\ \cline{2-5}
 & 5 & 54.5\% & [ 47.3\%, 61.6\% ] & 198 \\ \cline{2-5}
 & avg. \textbf{;} std & \multicolumn{1}{|c}{ 49.5\% } & \multicolumn{2}{@{}l|}{\textbf{;}\ \ \ 08.0\% } \\
\specialrule{0.8pt}{0pt}{0pt}
\multirow{6}{3em}{Front Leg(s)}  & 1 & 50.5\% & [ 43.7\%, 57.2\% ] & 220 \\ \cline{2-5}
 & 2 & 31.0\% & [ 24.7\%, 37.9\% ] & 200 \\ \cline{2-5}
 & 3 & 26.0\% & [ 20.1\%, 32.7\% ] & 200 \\ \cline{2-5}
 & 4 & 46.0\% & [ 38.9\%, 53.2\% ] & 200 \\ \cline{2-5}
 & 5 & 13.3\% & [ 08.7\%, 19.2\% ] & 180 \\ \cline{2-5}
 & avg. \textbf{;} std & \multicolumn{1}{|c}{ 33.4\% } & \multicolumn{2}{@{}l|}{\textbf{;}\ \ \ 13.5\% } \\
\specialrule{0.8pt}{0pt}{0pt}
\multirow{6}{3em}{Head}  & 1 & 55.7\% & [ 48.7\%, 62.5\% ] & 210 \\ \cline{2-5}
 & 2 & 44.8\% & [ 37.9\%, 51.8\% ] & 210 \\ \cline{2-5}
 & 3 & 58.1\% & [ 51.1\%, 64.8\% ] & 210 \\ \cline{2-5}
 & 4 & 41.9\% & [ 35.2\%, 48.9\% ] & 210 \\ \cline{2-5}
 & 5 & 65.8\% & [ 58.6\%, 72.5\% ] & 190 \\ \cline{2-5}
 & avg. \textbf{;} std & \multicolumn{1}{|c}{ 53.3\% } & \multicolumn{2}{@{}l|}{\textbf{;}\ \ \ 08.8\% } \\
\specialrule{0.8pt}{0pt}{0pt}
\multirow{6}{3em}{Tail}  & 1 & 57.4\% & [ 50.7\%, 63.9\% ] & 230 \\ \cline{2-5}
 & 2 & 46.4\% & [ 39.6\%, 53.2\% ] & 220 \\ \cline{2-5}
 & 3 & 46.4\% & [ 39.6\%, 53.2\% ] & 220 \\ \cline{2-5}
 & 4 & 10.5\% & [ 06.6\%, 15.6\% ] & 200 \\ \cline{2-5}
 & 5 & 18.6\% & [ 13.7\%, 24.4\% ] & 220 \\ \cline{2-5}
 & avg. \textbf{;} std & \multicolumn{1}{|c}{ 35.9\% } & \multicolumn{2}{@{}l|}{\textbf{;}\ \ \ 18.0\% } \\ \hline
\end{tabular}
\caption{\footnotesize Results from experiments determining the performance recognizing diagrammatic ASCII-art.
    The left-most column features the observed accuracy followed by an $\alpha=0.05$ Clopper-Pearson confidence bound
    on the true performance across the population.}
\label{table:apdx.moreRecognitionResults.humanArt1}
\end{table}
\normalsize

\newpage

\tiny
\begin{table}[h!]
\begin{tabular}{ |c|c|c|c|c| }
\hline
\multicolumn{5}{|c|}{Cars}\\ \hline
Part & Img Num. & Acc. & CI, $\alpha=0.05$ & Samp. Size \\ \hline
\multirow{5}{3em}{Body}  & 1 & 52.4\% & [ 44.6\%, 60.1\% ] & 170 \\ \cline{2-5}
 & 2 & 39.5\% & [ 33.0\%, 46.3\% ] & 220 \\ \cline{2-5}
 & 3 & 62.0\% & [ 54.9\%, 68.8\% ] & 200 \\ \cline{2-5}
 & 4 & 61.0\% & [ 54.0\%, 67.6\% ] & 210 \\ \cline{2-5}
 & avg. \textbf{;} std & \multicolumn{1}{|c}{ 53.7\% } & \multicolumn{2}{@{}l|}{\textbf{;}\ \ \ 09.0\% } \\
\specialrule{0.8pt}{0pt}{0pt}
\multirow{5}{3em}{Wheel(s)}  & 1 & 97.1\% & [ 93.9\%, 98.9\% ] & 210 \\ \cline{2-5}
 & 2 & 29.6\% & [ 23.7\%, 35.9\% ] & 230 \\ \cline{2-5}
 & 3 & 03.5\% & [ 01.4\%, 07.1\% ] & 200 \\ \cline{2-5}
 & 4 & 12.4\% & [ 08.2\%, 17.6\% ] & 210 \\ \cline{2-5}
 & avg. \textbf{;} std & \multicolumn{1}{|c}{ 35.6\% } & \multicolumn{2}{@{}l|}{\textbf{;}\ \ \ 36.7\% } \\
\specialrule{0.8pt}{0pt}{0pt}
\specialrule{0.4pt}{0pt}{0pt}
\multicolumn{5}{|c|}{Planes}\\ \hline
Part & Img Num. & Acc. & CI, $\alpha=0.05$ & Samp. Size \\ \hline
\multirow{11}{3em}{Tail}  & 1 & 51.9\% & [ 44.9\%, 58.8\% ] & 210 \\ \cline{2-5}
 & 2 & 62.4\% & [ 55.5\%, 69.0\% ] & 210 \\ \cline{2-5}
 & 3 & 35.9\% & [ 29.6\%, 42.6\% ] & 220 \\ \cline{2-5}
 & 4 & 21.4\% & [ 16.1\%, 27.4\% ] & 220 \\ \cline{2-5}
 & 5 & 26.2\% & [ 20.4\%, 32.7\% ] & 210 \\ \cline{2-5}
 & 6 & 57.6\% & [ 50.6\%, 64.4\% ] & 210 \\ \cline{2-5}
 & 7 & 15.9\% & [ 10.7\%, 22.3\% ] & 170 \\ \cline{2-5}
 & 8 & 31.0\% & [ 24.8\%, 37.7\% ] & 210 \\ \cline{2-5}
 & 9 & 41.8\% & [ 35.2\%, 48.6\% ] & 220 \\ \cline{2-5}
 & 10 & 63.3\% & [ 56.4\%, 69.9\% ] & 210 \\ \cline{2-5}
 & avg. \textbf{;} std & \multicolumn{1}{|c}{ 40.7\%} & \multicolumn{2}{@{}l|}{\textbf{;}\ \ \ 16.5\% } \\
\specialrule{0.8pt}{0pt}{0pt}
\multirow{11}{3em}{Wing(s)}  & 1 & 58.2\% & [ 51.4\%, 64.8\% ] & 220 \\ \cline{2-5}
 & 2 & 47.7\% & [ 41.0\%, 54.5\% ] & 220 \\ \cline{2-5}
 & 3 & 50.0\% & [ 42.9\%, 57.1\% ] & 200 \\ \cline{2-5}
 & 4 & 33.3\% & [ 27.0\%, 40.1\% ] & 210 \\ \cline{2-5}
 & 5 & 80.5\% & [ 74.2\%, 85.9\% ] & 190 \\ \cline{2-5}
 & 6 & 15.8\% & [ 10.9\%, 21.8\% ] & 190 \\ \cline{2-5}
 & 7 & 34.0\% & [ 27.5\%, 41.0\% ] & 200 \\ \cline{2-5}
 & 8 & 37.4\% & [ 30.5\%, 44.7\% ] & 190 \\ \cline{2-5}
 & 9 & 58.5\% & [ 51.3\%, 65.4\% ] & 200 \\ \cline{2-5}
 & 10 & 27.8\% & [ 21.4\%, 34.9\% ] & 180 \\ \cline{2-5}
 & avg. \textbf{;} std & \multicolumn{1}{|c}{44.3\%} & \multicolumn{2}{@{}l|}{\textbf{;}\ \ \ 17.6\% } \\ \hline
\end{tabular}
\caption{\footnotesize Results from experiments determining the performance recognizing diagrammatic ASCII-art.
    The left-most column features the observed accuracy followed by an $\alpha=0.05$ Clopper-Pearson confidence bound
    on the true performance across the population.}
\label{table:apdx.moreRecognitionResults.humanArt2}
\end{table}

\normalsize

\subsection{Example Airplane ASCII-Art As Found In Prompt And Queries to GPT3.5}
\label{subsubsec:humanAsciiArt.examplePlanes}

\begin{figure}
\begin{lstlisting}[basicstyle=\ttfamily\tiny,keepspaces=true,language={}]
Exemplar 6 from prompt:

   __|__
*---o0o---*

Airplane image 1:

       __|__
--@--@--(_)--@--@--

Airplane image 2:

       __|__
--o--o--(_)--o--o--

Airplane image 3:

      __|__
*---o--(_)--o---*

Airplane image 4:

    __!__
_____(_)_____
   !  !  !

Airplane image 5:

       __|__
--------(_)--------
  O  O       O  O

Airplane image 8:

 __
 \  \     _ _
  \**\ ___\/ \
X*#####*+^^\_\
  o/\  \
     \__\

Airplane image 10:

            __/\__
           `==/\==`
 ____________/__\____________
/____________________________\
  __||__||__/.--.\__||__||__
 /__|___|___( >< )___|___|__\
           _/`--`\_
          (/------\)

\end{lstlisting}
\caption{
\footnotesize Examples of ASCII-art used for airplanes. The top illustration
was included in the prompt, along with an indication of where the wings were. While
subtly different, we show images 1 through 5 as those that, at first glance, have noticable
similarity to the exemplar. Image 8 and 10 are included to demonstrate the diversity present
in the rest of the airplane images.}
\label{apdx:fig.humanASCIIArt.airplanes}
\normalsize
\end{figure}
\newpage

\section{Remarks Regarding a Subset %
    of Our Investigations Leading Up to this Work: Leveraging 
    Existing VQA Datasets And ASCII-Art Converters }
\label{appdx:OtherInvestigationsFromPast}

Leading up to the results presented in this paper, we explored
various flavors of visual interaction with ChatGPT and GPT3.5.
While we found most of these efforts enlightening in one way or another, 
we refrain (at this time, anyway) from detailing each branch we 
visited. In the text's main body we already commented on at least two
cases that helped shape the direction of our investigation.
We take a moment now, though, to discuss an early route we invested part of 
our time exploring: examination of how well ChatGPT could perform on 
existing, respected VQA tasks, using ASCII-art generators as the conduit between
pixel/vector graphics and the LLM's text input.

We investigated at least seven different ASCII-art generators, %
preferring those that had open-source code which we could use locally.
In the end, we found %
\cite{BizauGithub} 
to be the most workable tool. We continued with this tool somewhat reluctantly since it was a 
solid-fill/tone-based converter, a fact which tended to make structures pertinent to our tasks %
difficult to see at the sizes we were considering; we'll comment shortly on the additional efforts
we attempted to mitigate the difficulties.
Relatively late in the exploration of feeding ChatGPT existent VQA data,
we came across the line-art generating method discussed in \cite{DBLP:journals/tog/XuZW10}.
Though eager to try this approach, we were unable to find publicly-shared implementations that fully functioned %
for our purposes, even after reasonable attempts to get them running.
We explored other line-art generating methods, sanity-checking a few with online interfaces the 
authors provided, but found none that met our criteria better than the alternatives we already had working.
 
In regards to data we wanted to try converting, 
our early swings considered using images and --- importantly --- their corresponding questions
from the CLEVR VQA diagnostic dataset \cite{DBLP:conf/cvpr/JohnsonHMFZG17}.
We doubted ChatGPT would be capable of handling this challenge well, but we felt it was worth exploring, if nothing
else than to establish an upper-bound on ability.
Unfortunately, the three dimensional rendering, shadows, texturing, and reflections that make CLEVR a
reasonably good synthetic dataset to benchmark VQA systems on contributed to multiple difficulties in
generating ASCII-art representations that unmistakably reflected the objects in the depiction.
The character-art extracted tended to be 
too noisy, feature extra lines and shading from shadows and reflections,
and were not easy to interpret\footnote{``Easy to interpret'' as gauged by 
whether the family of questions one might ask about the image could be reliable answered
with the information shown.}
 at the resolutions we could use.
This persisted despite combinations of cropping, converting to gray-scale,
mild blurring, and performing Canny edge-detection \cite{DBLP:journals/pami/Canny86a}  prior to feeding material into the 
ASCII-art extractor.

Elaborating on resolution limits, %
the LLM's context window places a hard-limit on the maximum size of an ASCII-art depiction.\footnote{We suspect
that performance would be negatively impacted by increasing the size, all else equal. While we kept potential
soft-impacts in mind, the hard-limit and clarity issues more heavily dominated the adjustments we had to perform.} 
If we ignore the compressive effect of tokenization and suppose each token represents one letter,
a context window of 4096 tokens could hold only a 64-by-64 ASCII-art image, less if we count space for
newlines and a prompt. Our belief is that reasonable, line-like ASCII-art reliably achieves a higher 
token compression ratio than $1:1$, however the 64-by-64 cap is still a valid worse-case maximum to our knowledge.
We were in fact able to experiment with character-art at larger scales. Still, there exists a hard-cap that is comparatively 
small for an image, and moreover, we felt slight discomfort at the idea of pushing sizes to their maximum when 
performance is reputed to negatively correlate with relationship distance
for most
neural-network based NLP systems, including LLMs. %

Searching for an alternative for the CLEVR dataset that retained
its pros while addressing sources of difficulty, we came across 
Shapeworld \cite{kuhnle2017shapeworld}, a synthetic VQA dataset generator that produced 2-D 
images filled with geometric entities 
of uniform texture and flat color that did not have shadows or reflections.
We modified the code associated with the project to keep
only the shapes that seemed mutually differentiable and/or individually discernible in the 
resolution of ASCII-art we could provide --- this left
squares, rectangles, triangles, circles and crosses, and dropped pentagons, semicircles, and 
ellipses. 
The collision tolerance was reduced to zero, and the maximum number of shapes we allowed was four.
We initially tried to retain the diversity of color among individuals: the ASCII-art generator we
used was capable of outputting 
text that, via computer terminal encodings, would be rendered with that additional information displayed.
Ultimately, the presence of multiple colors was not viable, due to the extra space the encoding took. 
Additionally, %
we walked away with the 
impression that the shell encodings distracted the network from the shapes that were being presented;
while a more rigorous analysis would be needed to assert that with confidence, we at least
got the sense that multi-color was not helpful beyond enabling minimally interesting questions.
As an example of an uninteresting question multi-coloring facilitated, consider
``Is there something green?'', which could be answered by looking up which shell encoding corresponds to green
and whether that substring is present, in contrast to a query like ``Is there a green square to the left of a blue circle?''
that would require a deeper appreciation of the scene while drawing multiple facets of the image together.
In the end, we required Shapeworld to produce all figures in gray. 
From there, we passed the resultant images through Canny edge-detection prior to forwarding 
them to the ASCII-art converter.
Corresponding vocabulary available to Shapeworld also had to be stripped to remove references to the 
options we decided against leveraging. %

We of course examined use of the ASCII-art converter for images not native to a pre-existing VQA dataset.
For instance, using line-drawings we manually produced in an electronic power-point program then export as PNGs, 
we were able to produce %
ASCII depictions of comparatively high fidelity.
As weighed against the prior paragraphs, however,
these efforts had the downside of %
not already having questions available which meaningfully related to the images, nor did they have 
corresponding answers. 
Additionally, the prior VQA datasets have the advantage of being more thoroughly studied and being 
a known-quantity, facts that perhaps would have allowed more knowledge to come to bear during analysis 
as well as to reduce our explanatory burden.

Across these attempts, we did not find ChatGPT's performance on these exploratory questions to be 
sufficiently compelling as to divert more attention in this direction %
at the cost of alternatives we
were concurrently investigating. Since the time of these attempts, however, we have progressively refined
our processes for interacting with the LLM, %
so it is possible that future work may take another crack at this, 
if nothing else than to more rigorously confirm that the task is relatively difficult for ChatGPT.

\begin{figure}[h!]
\begin{tabular}{c}
\includegraphics[width=0.3\linewidth]{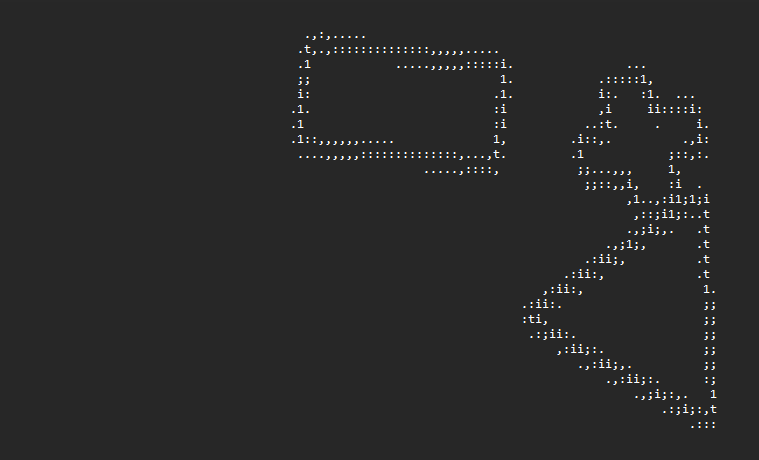}
\includegraphics[width=0.3\linewidth]{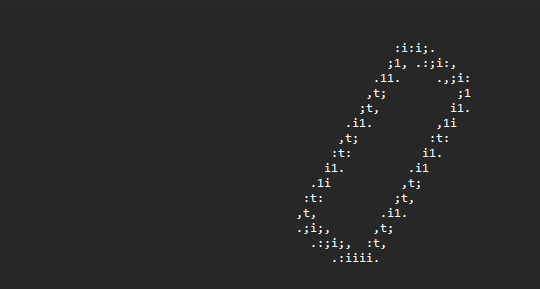}
\end{tabular}
\caption{\tiny Two examples of ASCII-art produced by converting images from 
     Shapeworld after we restricted the set of output it was allowed to produce and ran
    the results through Canny edge detection.}
\label{fig:shapeworldAsAsciiLineArt}
\end{figure}